\PassOptionsToPackage{table}{xcolor}
\documentclass[10pt,twocolumn,letterpaper]{article}
\usepackage[T1]{fontenc}  

\usepackage[pagenumbers]{cvpr} 

\usepackage{pifont}
\usepackage{caption}
\usepackage{booktabs}
\usepackage{multirow}
\usepackage[table]{xcolor}
\usepackage{graphicx}
\usepackage{fontawesome5}

\definecolor{gtgray}{RGB}{150,150,150}
\definecolor{oursblue}{RGB}{65,115,210}
\definecolor{tramred}{RGB}{210,55,55}
\definecolor{rolloutorange}{RGB}{217,105,42}
\definecolor{figblue}{RGB}{36,95,208}

\newcommand{\colorswatch}[1]{%
    \textcolor{#1}{\rule{0.65em}{0.65em}}}

\definecolor{cvprblue}{rgb}{0.21,0.49,0.74}
\usepackage[pagebackref,breaklinks,colorlinks,allcolors=cvprblue]{hyperref}

\def\confName{CVPR}
\def\confYear{2027}

\title{GravHMR: Coupling SLAM and Human Mesh Recovery in a Shared Gravity Frame}

\newcommand{\afflogo}[2][1.55em]{\raisebox{-0.32\height}{\includegraphics[height=#1]{#2}}\hspace{0.45em}}
\makeatletter
\@ifundefined{iflatexml}{\IfFileExists{latexml.sty}{\usepackage{latexml}}{\newif\iflatexml}}{}
\makeatother
\iflatexml
\author{
Hanglei Jin\textsuperscript{1,2,*,\S} \and
Chenglu Sun\textsuperscript{2,*,\ddag} \and
Junchen Ge\textsuperscript{1} \and
Yuchao Wang\textsuperscript{2} \and
Siqi Chen\textsuperscript{1} \and
Yuzhi Huang\textsuperscript{1} \and
Jingyan Jiang\textsuperscript{3} \and
Zhi Wang\textsuperscript{1,\dag}
\and
\textsuperscript{1}Shenzhen International Graduate School, Tsinghua University;
\textsuperscript{2}Tencent D-Live;
\textsuperscript{3}Shenzhen Technology University
\and
\textsuperscript{*}Equal contribution;
\textsuperscript{\ddag}Project lead;
\textsuperscript{\dag}Corresponding author;
\textsuperscript{\S}Work done during internship at Tencent
\and
Project page: \href{https://jhl25-pixel.github.io/gravhmr/}{jhl25-pixel.github.io/gravhmr}
}
\else
\author{
Hanglei Jin\textsuperscript{1,2,*,$\S$} \quad
Chenglu Sun\textsuperscript{2,*,$\ddagger$} \quad
Junchen Ge\textsuperscript{1} \quad
Yuchao Wang\textsuperscript{2} \\[3pt]
Siqi Chen\textsuperscript{1} \quad
Yuzhi Huang\textsuperscript{1} \quad
Jingyan Jiang\textsuperscript{3} \quad
Zhi Wang\textsuperscript{1,$\dagger$}\\[9pt]
\normalsize\afflogo{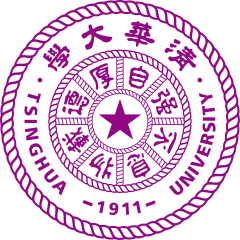}\textsuperscript{1}Shenzhen International Graduate School, Tsinghua University\\[4pt]
\normalsize\afflogo[1.45em]{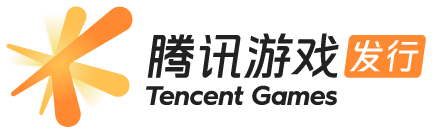}\textsuperscript{2}Tencent D-Live
\hspace{1.6em}\afflogo[1.45em]{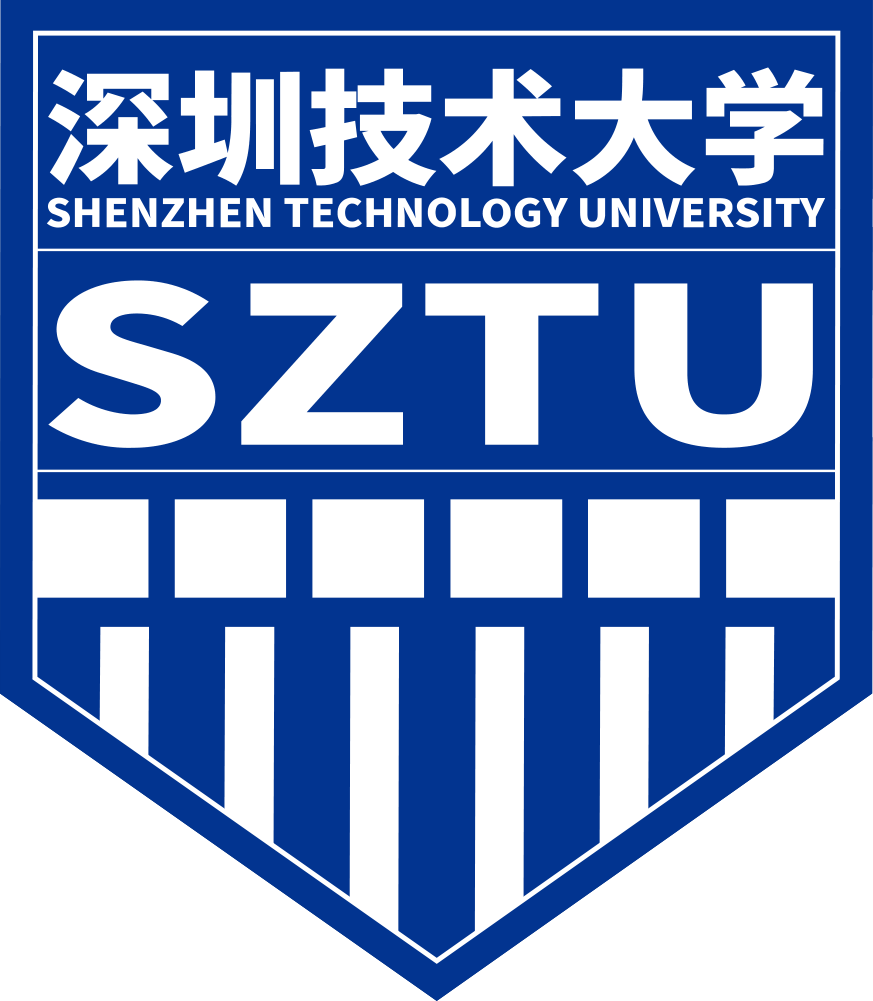}\textsuperscript{3}Shenzhen Technology University\\[6pt]
\small\textsuperscript{*}Equal contribution \quad
\textsuperscript{$\ddagger$}Project lead \quad
\textsuperscript{$\dagger$}Corresponding author \quad
\textsuperscript{$\S$}Work done during internship at Tencent\\[5pt]
\normalsize{\faGithub}\hspace{0.35em}Project page: \href{https://jhl25-pixel.github.io/gravhmr/}{jhl25-pixel.github.io/gravhmr}
}
\fi

\begin{document}

\maketitle

\begin{abstract}
Recovering globally consistent 3D human motion from
monocular videos remains challenging under dynamic
camera motion. Although SLAM and human mesh recovery
(HMR) provide complementary camera and body estimates,
directly combining them often leads to inconsistent
orientations and accumulated trajectory drift.
We present \textbf{GravHMR}, a training-free, plug-and-play framework that couples SLAM and HMR through a shared gravity-aligned reference. By allowing gravity to be estimated independently of the HMR backbone, GravHMR accommodates world-grounded HMR models with or without explicit gravity-view representations. Specifically, Yaw-Aware Orientation Optimization (YAOO) resolves heading inconsistencies by correcting body and velocity orientations about the camera-frame gravity axis. Conditioned on these corrected orientations, Gravity-Aware Trajectory Optimization (GATO) anchors root trajectories in the metric SLAM world by jointly leveraging 2D reprojection, the metric scale implicit in the human body, and the body's own motion.
On EMDB-2, GravHMR achieves state-of-the-art performance
with 154.0\,mm W-MPJPE$_{100}$ using ViPE and 189.8\,mm using masked DROID-SLAM,
outperforming TRAM (222.4\,mm) under the same masked DROID-SLAM camera estimates.
It further reduces W-MPJPE$_{100}$ by 44.3\% over GVHMR (276.5\,mm) and by 20.0\% over HTD-Refine (192.4\,mm) with the same HMR backbone.

\end{abstract}

\section{Introduction}
\label{sec:intro}

\begin{figure}[t]
  \centering
  \includegraphics[width=\linewidth]{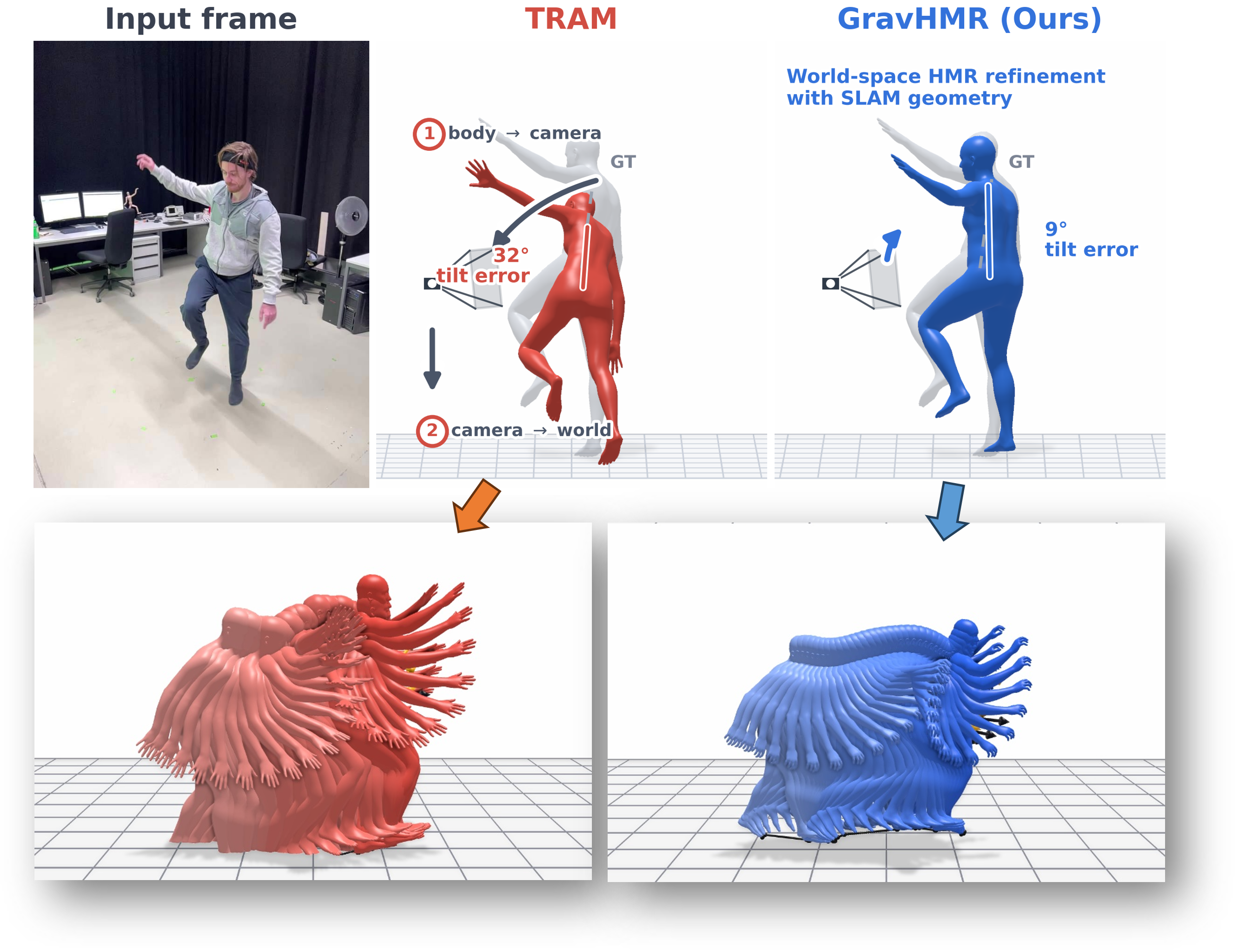}
  \caption{\textbf{Motivating observation: orientation inconsistency in world-grounded human reconstruction.}
  Directly lifting camera-space body estimates into the world using independently estimated SLAM camera poses can propagate orientation errors, as illustrated by TRAM (\textcolor{tramred}{red}).
  In contrast, GravHMR reconciles body and camera orientations through a shared gravity reference, yielding an upright reconstruction more consistent with the ground truth (\textcolor{gtgray}{gray}).}
  \label{fig:teaser}
\end{figure}

Recovering globally consistent 3D human motion
from monocular videos requires reconciling two
complementary sources of information: human
motion priors and camera geometry. Modern
human mesh recovery (HMR) models capture
articulated body motion and provide valuable
orientation and motion priors, while SLAM
estimates temporally consistent camera poses
in a metric world. Despite their complementary
strengths, combining independently estimated
camera and human motion remains challenging.
A geometrically valid camera-to-world
transformation does not necessarily yield
a physically consistent human reconstruction:
discrepancies between the camera geometry
and the HMR motion reference can manifest
as incorrect body headings and drifting
global trajectories.

The central challenge is therefore not merely
to refine rotation and translation independently,
but to determine how complementary geometric
information should be transferred between
SLAM and HMR. Directly adopting the full
camera-derived body orientation may discard
useful gravity-referenced tilt information
from HMR models such as GVHMR, whereas relying solely on the
predicted human-motion rollout can retain
inconsistent headings and accumulated drift.
Moreover, these discrepancies interact:
an incorrect heading biases the motion
prior and geometric constraints used for
root trajectory recovery. This motivates a selective and ordered coupling
strategy: first correcting the horizontal headings
of the body and its motion about the shared gravity
axis, and then recovering root displacements
under the corrected body orientations.

Our main contributions are:
\begin{itemize}
    \item We introduce \textbf{GravHMR}, a training-free
    framework that couples independently estimated
    SLAM camera geometry and HMR motion priors
    through a shared gravity reference, enabling
    consistent world-grounded human motion recovery
    without retraining the underlying HMR backbone.

    \item We propose a gravity-constrained geometric
    coupling strategy with two complementary modules:
    Yaw-Aware Orientation Optimization (YAOO) and
    Gravity-Aware Trajectory Optimization (GATO).
    YAOO reconciles body and motion orientations
    within the shared gravity reference for \textbf{better human orientation}, while GATO
    anchors root trajectories in the metric SLAM world
    using geometric observations and gravity-referenced
    motion priors.
    Together, they improve body tilt and vertical
    alignment, travel direction and human root trajectory.

    \item GravHMR improves the global trajectories
    of its HMR backbone.
    On EMDB-2, it achieves
    154.0\,mm W-MPJPE$_{100}$ using ViPE and
    189.8\,mm using masked DROID-SLAM,
    reducing global motion error by 44.3\%
    relative to the reported GVHMR baseline
    and by 20.0\% relative to GVHMR + HTD-Refine;
    on RICH, it reduces the error by 11.9\%
    relative to GVHMR + HTD-Refine.
    Under shared masked DROID-SLAM camera estimates,
    it reduces the error by 14.7\% over TRAM
    and 7.4\% over TRAM + HTD-Refine.
\end{itemize}
\section{Related Work}
\label{sec:related_work}

\subsection{Monocular Human Motion Recovery and Pose Estimation}

Monocular human motion recovery builds on parametric
human body models~\cite{loper2023smpl,SMPLX2019} to
estimate pose and shape from images.                                     
HMR~\cite{kanazawa2018end} directly regresses body
parameters from a single image with an adversarial prior.
SMPLify~\cite{SMPLify2016} instead fits SMPL to 2D
keypoints through iterative optimization.                                
SPIN~\cite{kolotouros2019learning} integrates model
fitting into the training loop to improve regression
through optimized body estimates.
CLIFF~\cite{li2022cliff} incorporates bounding-box
location and full-frame reprojection supervision to
improve camera-relative orientation and pose estimation,
while SPEC~\cite{SPEC2021} and CameraHMR~\cite{CameraHMR2024}
further condition the regression on estimated camera
intrinsics.                                                              
Part-aware designs such as PARE~\cite{PARE2021} and
analytical inverse-kinematics models such as
HybrIK~\cite{HybrIK2021} improve robustness under
occlusion and extreme poses.                                             
HMR~2.0~\cite{goel2023humans} adopts a vision
transformer and a transformer decoder for robust
single-image human mesh recovery, and ReFit~\cite{ReFit2023}
iteratively refines its predictions through recurrent
feedback.                                                                
Multi-HMR~\cite{baradel2024multi} extends feed-forward
recovery to multiple whole-body meshes in a single
image.
PromptHMR~\cite{wang2025prompthmr} conditions
full-image mesh recovery on spatial and semantic
prompts, supporting partial observations and
person--person interactions.

For video inputs, temporal modeling provides
additional evidence for coherent motion reconstruction.
VIBE~\cite{kocabas2020vibe} combines temporal
regression with an adversarial motion discriminator
trained using unpaired motion-capture data.
TCMR~\cite{choi2021beyond} emphasizes temporal context
to improve motion consistency while preserving
per-frame accuracy.
HTD-Refine~\cite{wei2026natural} estimates
image-conditioned 3D joint velocities and accelerations
and uses them to refine existing motion reconstructions.
These advances improve body reconstruction and temporal
dynamics, while recovering global trajectories under
moving cameras additionally requires resolving
camera--human motion ambiguity.

\subsection{Global Human Motion Recovery from Moving-Camera Videos}

Recovering world-space human motion requires separating
human movement from camera ego-motion, which in turn
relies on robust visual SLAM such as
DROID-SLAM~\cite{teed2021droid} and
DPVO~\cite{DPVO2023} for camera-pose initialization.    
GLAMR~\cite{yuan2022glamr} predicts global human
trajectories from local body motion and jointly
optimizes human trajectories and camera poses against
image observations.
TRACE~\cite{TRACE2023} regresses world-space 5D
trajectories directly from in-the-wild video.          
SLAHMR~\cite{ye2023decoupling} combines SLAM-derived
camera motion with learned human motion priors to
resolve scale ambiguity and recover global human
trajectories.
PACE~\cite{kocabas2024pace} tightly couples human
motion priors and background scene features in a
bundle-adjustment-inspired optimization of human and
camera motion.
WHAM~\cite{shin2024wham} combines human motion
features with SLAM-estimated camera angular velocity
to recover world-grounded motion.
GVHMR~\cite{shen2024world} predicts body orientation
in gravity-view coordinates and combines it with
camera rotation and local root velocity to reconstruct
global motion.
TRAM~\cite{wang2024tram} combines masked SLAM and
background-derived metric scale with a video-based
human motion regressor to recover world-space
trajectories, and GENMO~\cite{li2025genmo} further
casts global motion recovery as a conditional
generative problem over a large motion prior.          
PromptHMR-Vid~\cite{wang2025prompthmr} extends
promptable mesh recovery to video and combines its
predictions with TRAM's metric SLAM for global motion
estimation.
Humans as Checkerboards~\cite{yang2025humans} uses
HMR-derived human--scene contact depths to calibrate
monocular SLAM scale without iterative optimization,
and \cite{choi2025humans} similarly treats the human
body as a dynamic calibration pattern for
multi-camera scene reconstruction.                     
Joint human--scene optimization has also been
explored by JOSH~\cite{liu2025joint} and
HumanMM~\cite{HumanMM2025} for multi-shot videos.      

Despite these advances, independently estimated
camera geometry and human motion predictions can
remain inconsistent.
Integrating predicted root velocities may accumulate
trajectory drift, while camera-space lifting alone
does not explicitly reconcile image observations with
the temporal motion prior.
GravHMR addresses this inconsistency through
decoupled orientation and trajectory recovery:
camera geometry provides the world-space reference,
2D pose observations anchor root positions, and
gravity-referenced displacements constrain motion
consistency.

\section{GravHMR}
\label{sec:methods}


\begin{figure*}[t]
    \centering
    \includegraphics[width=0.9\textwidth]{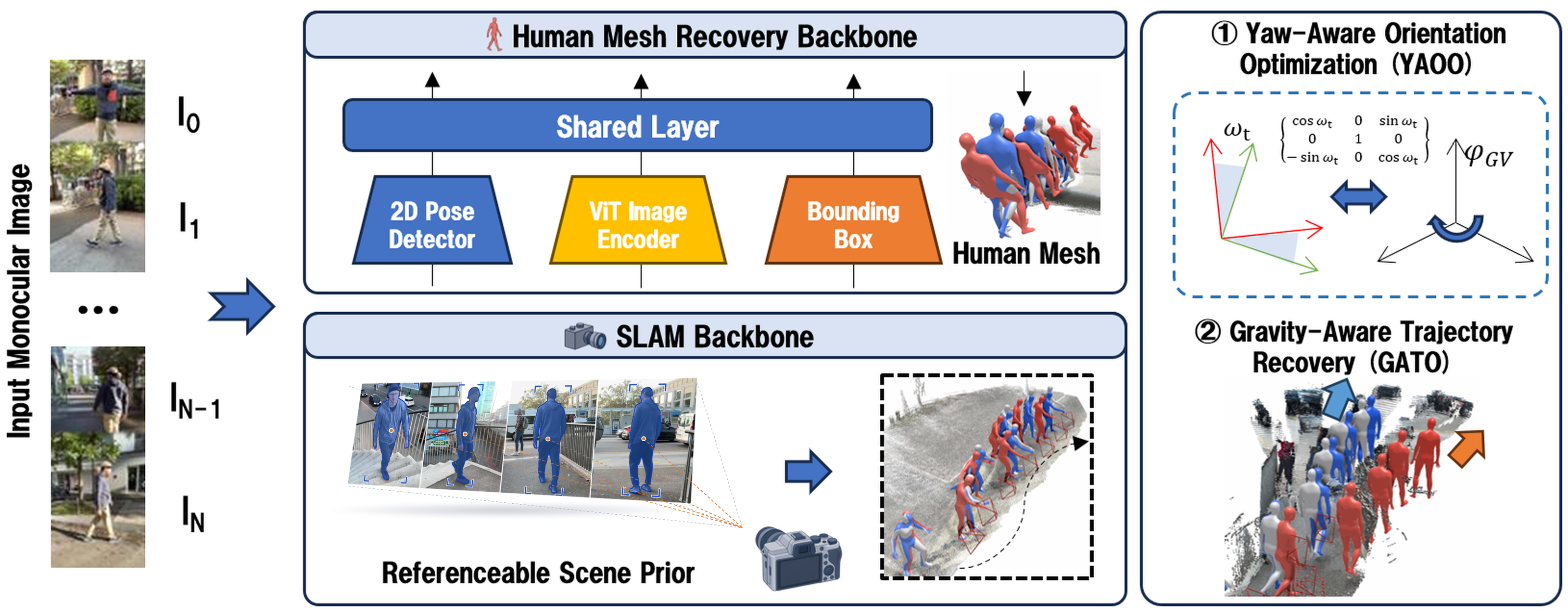}
    \caption{\textbf{Overview of GravHMR.}
    Given a monocular video, off-the-shelf HMR and SLAM
    backbones provide initial human motion and camera
    geometry, respectively.
    GravHMR couples these estimates through a shared
    gravity-aligned reference.
    Yaw-Aware Orientation Optimization (YAOO) corrects
    the headings of the body and of its motion about gravity
    while keeping the gravity-relative body tilt predicted by HMR.
    Gravity-Aware Trajectory Optimization (GATO) then recovers
    the global root trajectory from 2D pose observations
    under the corrected orientations and the SLAM camera geometry,
    and Flow-Aware Temporal Consistency (FATC) finally
    regularizes its temporal derivatives.}
    \label{fig:pipeline}
\end{figure*}

\subsection{Gravity-Aligned SLAM-HMR System}
\label{sec:gv_slam_hmr}

\noindent\textbf{SLAM and HMR Initialization.}
Given a monocular video $\{I_t\}_{t=1}^{T}$,
we use ViPE~\cite{huang2025vipe} and GVHMR~\cite{shen2024world}
to initialize camera and human motion, respectively.
ViPE initializes camera intrinsics using
GeoCalib~\cite{veicht2024geocalib} and subsequently refines
them jointly with camera poses and scene depth
through bundle adjustment.
We use its final camera intrinsics $\mathbf{K}_t$,
camera rotations $\mathbf{R}^{c\to w}_t$,
and camera centers $\mathbf{C}_t^{w}$.
The HMR backbone provides body pose, shape,
and camera-space root orientation and translation.

\noindent\textbf{Human in Scene with Gravity Direction.}
We obtain camera-frame gravity estimates
$\mathbf{g}^{c}_t$ from GeoCalib,
independently of the HMR backbone.
These estimates are used to construct our
gravity reference, rather than as explicit
constraints in ViPE's camera-pose optimization.
We transform the gravity estimates into the
SLAM world using the final camera rotations
and aggregate them into a shared vertical axis:
\begin{equation}
\mathbf{n}^{w}
=
-\frac{
\sum_{t=1}^{T}
\mathbf{R}^{c\to w}_t\mathbf{g}^{c}_t
}{
\left\|
\sum_{t=1}^{T}
\mathbf{R}^{c\to w}_t\mathbf{g}^{c}_t
\right\|_2
}.
\label{eq:gravity_axis}
\end{equation}
We then rotate the SLAM world by a rotation $\mathbf{R}^{w\to g}$ that
maps $\mathbf{n}^{w}$ to the $y$-axis; the heading about this axis is a
free gauge and does not affect the results. The cameras become
$\mathbf{R}^{c\to g}_t=\mathbf{R}^{w\to g}\mathbf{R}^{c\to w}_t$ and
$\mathbf{C}^{g}_t=\mathbf{R}^{w\to g}\mathbf{C}^{w}_t$, with
$\mathbf{R}^{g\to c}_t=(\mathbf{R}^{c\to g}_t)^{\top}$, and gravity
points along $\mathbf{n}^{g}=[0,1,0]^\top$.
Because gravity comes from the images rather than from the
HMR model, this step works for HMR backbones with or without
a gravity-view representation.


\subsection{Human Root Trajectory Recovery}
\label{sec:root_recovery}

Given the SLAM camera parameters
$\{\mathbf{R}^{c\to g}_t,\mathbf{C}^{g}_t,\mathbf{K}_t\}$,
expressed in the gravity frame of Sec.~\ref{sec:gv_slam_hmr},
and the HMR outputs, we recover the world-space root trajectory
in two stages.
Yaw-Aware Orientation Optimization
(YAOO, Sec.~\ref{sec:yaoo}) resolves the headings of the body
and its motion in the shared gravity frame;
Gravity-Aware Trajectory Optimization (GATO, Sec.~\ref{sec:GATO})
then recovers the root trajectory under these orientations.

\subsubsection{Yaw-Aware Orientation Optimization}
\label{sec:yaoo}

The two sources of body orientation are reliable in different
components.
The HMR backbone estimates the body orientation relative
to the camera, $\mathbf{R}(\boldsymbol{\Gamma}^c_t)$,
far more reliably than the heading of its world-frame
rollout, $\mathbf{R}^{\mathrm{roll}}_t$---the gravity-view
body orientation produced by integrating the HMR motion
prior forward in time, whose heading about the gravity
axis drifts over time.
Conversely, the gravity-view rollout predicts the body tilt
directly with respect to gravity, whereas lifting the camera-frame
orientation into the world compounds the tilt errors of the
camera rotation and of the camera-frame prediction.
YAOO therefore takes from each source only what it estimates well:
the heading from the camera and the tilt from the rollout.

Concretely, we first lift the camera-frame orientation into
the gravity frame,
\begin{equation}
    \widetilde{\mathbf{R}}_t
    \;=\;
    \mathbf{R}^{c\to g}_t\,
    \mathbf{R}(\boldsymbol{\Gamma}^c_t),
    \label{eq:yaoo_ref}
\end{equation}
where $\boldsymbol{\Gamma}^c_t$ is the HMR camera-frame root
orientation and $\mathbf{R}(\cdot)$ its rotation matrix,
and extract its heading relative to the rollout as the
rotation about the gravity axis $\mathbf{n}^{g}$ that
aligns $\mathbf{R}^{\mathrm{roll}}_t$ with
$\widetilde{\mathbf{R}}_t$,
\begin{equation}
    \psi_t
    \;=\;
    \mathrm{Yaw}_{\mathbf{n}^{g}}\!
    \big(\,\widetilde{\mathbf{R}}_t
    (\mathbf{R}^{\mathrm{roll}}_t)^{\top}\,\big),
    \label{eq:yaoo_psi}
\end{equation}
where $\mathrm{Yaw}_{\mathbf{n}^{g}}(\mathbf{R})$ is the signed
angle of $\mathbf{R}$ about $\mathbf{n}^{g}$, \ie the projection
of its axis--angle vector onto $\mathbf{n}^{g}$.
We apply only this yaw to the rollout and transfer the
same correction to the rollout displacement:
\begin{equation}
\begin{aligned}
    \mathbf{R}_t
    &= \exp\!\big(\psi_t\,[\mathbf{n}^{g}]_\times\big)\,
       \mathbf{R}^{\mathrm{roll}}_t,\\
    \mathbf{d}_t
    &= \exp\!\big(\psi_t\,[\mathbf{n}^{g}]_\times\big)\,
       \big(\mathbf{X}^{\mathrm{roll}}_{t+1}
            - \mathbf{X}^{\mathrm{roll}}_t\big),
\end{aligned}
\label{eq:yaoo_apply}
\end{equation}
where $[\cdot]_\times$ is the skew-symmetric matrix and
$\mathbf{X}^{\mathrm{roll}}_t$ the rollout root position.
Eq.~\eqref{eq:yaoo_apply} is a pure rotation about the gravity
axis: the heading of $\mathbf{R}_t$ follows the camera-anchored
reference of Eq.~\eqref{eq:yaoo_ref}, which does not accumulate
drift, while its gravity-relative tilt remains that of the rollout.
Adopting $\widetilde{\mathbf{R}}_t$ in full would instead also
import its tilt, which reduces accuracy
(Sec.~\ref{sec:ablation}, Tab.~\ref{tab:orient_src}).
Each rollout displacement is rotated by the same angle $\psi_t$,
which preserves its magnitude and the gait predicted by HMR and
changes only the horizontal direction in which it is applied.
The corrected displacement $\mathbf{d}_t$ is not integrated
into a trajectory; it serves as a heading-consistent motion
prior for GATO.

\subsubsection{Gravity-Aware Trajectory Optimization}
\label{sec:GATO}

\paragraph{Root Initialization.}
A 2D keypoint constrains the root only up to its depth along
the viewing ray.
For each frame, we fit the camera-frame root translation to the
confidence-weighted 2D keypoints under the camera-frame body
orientation $\mathbf{R}^{g\to c}_t\mathbf{R}_t$; we refer to this
as the per-frame 2D root solve.
Its viewing direction $\mathbf{r}_t$, normalized such that
$[\mathbf{r}_t]_z=1$, is well constrained, but its depth is set only
by the apparent body size---a body of metric size $H_{\mathrm{body}}$
spanning $h_t$ pixels under focal length $f_t$ lies at
$z_t\approx f_t H_{\mathrm{body}}/h_t$---and becomes unreliable once
the body is truncated.
We therefore take the depth from the HMR camera-space root
translation, $z^{\mathrm{HMR}}_t$, which the network regresses from
the image even when the body is partly out of view, and initialize
the world-space root as
\begin{equation}
    \mathbf{X}^{0}_t
    = \mathbf{C}^{g}_t
    + z^{\mathrm{HMR}}_t\,
      \mathbf{R}^{c\to g}_t\,\mathbf{r}_t.
    \label{eq:init}
\end{equation}

\paragraph{Root Trajectory with Self-Calibrated Weighting.}
With $\{\mathbf{R}_t\}$ fixed by YAOO and initialized from
Eq.~\eqref{eq:init}, we refine the per-frame root positions
$\mathcal{X}=\{\mathbf{X}_t\}_{t=1}^{T}$.
The image alone leaves two problems: a keypoint fixes the root
only up to its depth along the viewing ray, and the visible
keypoints change from frame to frame under occlusion and
truncation. GATO resolves the depth with the human body and
bridges the weakly observed frames with the body's own motion:
\begin{equation}
\mathcal{X}^{*} \;=\; \arg\min_{\mathcal{X}}\;
\mathcal{L}_{\text{reproj}}
\,+\, \mathcal{L}_{\text{human-scale}}
\,+\, \mathcal{L}_{\text{dynamic}}.
\label{eq:GATO}
\end{equation}
Each term applies the Cauchy penalty
$\rho_{\sigma}(r)=\log(1+r^2/\sigma^2)$ to its residuals, and the
three terms are summed without weights.
Each scale $\sigma$ is self-calibrated from the input by the
median absolute deviation against the per-frame 2D root solve,
without ground truth (Appendix~\ref{appendix:sigma}).

\noindent\textit{Reprojection.}
Under the metric SLAM camera, each detected keypoint
$\mathbf{u}_{t,k}$ with detection confidence $c_{t,k}$ constrains
the root through the root-relative joint
$\mathbf{J}_k(\boldsymbol{\theta}_t,\boldsymbol{\beta})$:
\begin{equation}
\mathcal{L}_{\text{reproj}}
=
\sum_{t,k} c_{t,k}\,
\rho_{\sigma_{\mathrm{2D}}}\!\Big(
\big\|\,\Pi_t(\mathbf{R}_t\mathbf{J}_k+\mathbf{X}_t)
       -\mathbf{u}_{t,k}\,\big\|_2\Big),
\label{eq:Lreproj}
\end{equation}
where $\Pi_t(\mathbf{X})=\pi\big(\mathbf{K}_t\mathbf{R}^{g\to c}_t
(\mathbf{X}-\mathbf{C}^{g}_t)\big)$ projects with the SLAM camera of
frame $t$, $\pi(\cdot)$ is the perspective division, and
$\boldsymbol{\theta}_t$, $\boldsymbol{\beta}$ are the HMR body pose
and shape.
The confidence weights let occluded or truncated joints contribute
little (Appendix~\ref{appendix:jointweight} compares per-joint
weightings).

\noindent\textit{Human scale.}
Reprojection leaves the root free along its viewing ray.
The HMR depth $z^{\mathrm{HMR}}_t$ fixes where on the ray the
root lies:
\begin{equation}
\mathcal{L}_{\text{human-scale}}
=
\sum_{t}
\rho_{\sigma_{z}}\!\Big(
\big[\mathbf{R}^{g\to c}_t(\mathbf{X}_t-\mathbf{C}^{g}_t)\big]_z
       - z^{\mathrm{HMR}}_t\Big),
\label{eq:Lhs}
\end{equation}
where $[\cdot]_z$ takes the depth component.

\noindent\textit{Dynamic.}
When joints are occluded or leave the image, their weights in
Eq.~\eqref{eq:Lreproj} drop, and a frame may keep too few
keypoints to locate the root. The body's motion is still
available: the HMR rollout predicts a locally consistent
displacement $\mathbf{d}_t$. We penalize the deviation of the
root displacement from $\mathbf{d}_t$:
\begin{equation}
\mathcal{L}_{\text{dynamic}}
=
\sum_{t=1}^{T-1}
\rho_{\sigma_{d}}\!\Big(
\big\|\,(\mathbf{X}_{t+1}-\mathbf{X}_t)
       - \mathbf{d}_t\,\big\|_2\Big).
\label{eq:Ldyn}
\end{equation}
On weakly observed frames, this term carries the root from
its neighbors. It also adapts to the motion: on static frames,
$\mathbf{d}_t\!\to\!0$ holds the root in place, and on fast
motion, $\mathbf{d}_t$ lets the trajectory follow the large
inter-frame displacement.

\subsection{Temporal Consistency Recovery}
\label{sec:htd}

GATO and higher-order motion cues are complementary in frequency.
GATO anchors the root to absolute positions through per-frame
reprojection, which removes low-frequency heading and trajectory
drift but leaves each frame weakly coupled to its neighbors,
so residual high-frequency jitter can remain.
Image-derived kinematic cues show the opposite behavior.
The joint velocities and accelerations predicted by
PVA-Net~\cite{wei2026natural} are locally accurate and temporally
smooth, but integrating them accumulates drift and therefore
cannot determine the global trajectory on their own.

Since GATO already provides a drift-free trajectory, these cues
no longer need to recover global position; they only need to
regularize local temporal derivatives around a well-anchored
solution.
We therefore add a Flow-Aware Temporal Consistency (FATC) stage
that reuses the refinement formulation of
HTD-Refine~\cite{wei2026natural}: starting from the GATO
reconstruction, we jointly refine body pose and root motion so
that their temporal derivatives agree with the velocities and
accelerations predicted by PVA-Net.
FATC is not designed to recover the global trajectory; it inherits
the trajectory of GATO and mainly removes high-frequency jitter and
foot sliding. Sec.~\ref{sec:exp_world} reports its effect separately.


\section{Experiments}
\label{sec:experiments}



\definecolor{blockrow}{RGB}{232,238,247}
\definecolor{bestcell}{RGB}{210,228,242}
\definecolor{secondcell}{RGB}{235,243,249}

\newcommand{\rgbrank}[2]{#1}

\newcommand{\rgbrankB}[2]{%
  \cellcolor{bestcell}\textbf{#1}}

\newcommand{\rgbrankU}[2]{%
  \cellcolor{secondcell}\underline{#1}}

\begin{table*}[t]
\centering
\small
\caption{\textbf{Global human motion on EMDB-2 and RICH.}
WA-/W-MPJPE$_{100}$ and FS in mm, RTE in \%, Jitter in $10\,\mathrm{m/s^3}$.
\emph{Ours w/ ViPE} uses ViPE camera poses and estimated intrinsics;
\emph{Ours w/ masked DROID-SLAM} uses the camera poses of TRAM and
ground-truth intrinsics, as TRAM does.
Best in \textbf{bold}, second \underline{underlined}.}
\label{tab:world}
\setlength{\tabcolsep}{3.5pt}
\renewcommand{\arraystretch}{1.08}
\resizebox{0.93\textwidth}{!}{%
\begin{tabular}{l ccccc ccccc}
\toprule
\multirow{2}{*}{Method}
& \multicolumn{5}{c}{EMDB-2}
& \multicolumn{5}{c}{RICH} \\
\cmidrule(lr){2-6}
\cmidrule(lr){7-11}
& WA-MPJPE$_{100}\downarrow$
& W-MPJPE$_{100}\downarrow$
& RTE$\downarrow$
& Jitter$\downarrow$
& FS$\downarrow$
& WA-MPJPE$_{100}\downarrow$
& W-MPJPE$_{100}\downarrow$
& RTE$\downarrow$
& Jitter$\downarrow$
& FS$\downarrow$ \\
\midrule
SLAHMR~\cite{ye2023decoupling}
& 326.9 & 776.1 & 10.2 & -- & --
& 98.1 & 186.4 & 28.9 & 34.3 & 5.1 \\
WHAM~\cite{shin2024wham}
& 135.6 & 354.8 & 6.0 & 22.5 & 4.4
& 109.9 & 184.6 & 4.1 & 19.7 & 3.3 \\
TRAM~\cite{wang2024tram}
& 76.4 & 222.4 & 1.4 & 18.5 & 23.4
& 103.6 & 168.4 & 2.7 & 18.7 & 12.9 \\
GVHMR~\cite{shen2024world}
& 111.0 & 276.5 & 2.0 & 16.7
& \rgbrankU{3.5}{100}
& 78.8 & 126.3 & 2.4 & 12.8
& \rgbrankU{3.0}{100} \\
GVHMR + PVA-Net~\cite{shen2024world,wei2026natural}
& 106.7 & 262.8 & 2.1 & 15.2
& \rgbrankB{3.4}{100}
& \rgbrankU{74.0}{100}
& \rgbrankU{119.1}{100}
& \rgbrankU{2.3}{100}
& 10.7 & \rgbrankB{2.7}{100} \\
PromptHMR-Vid~\cite{wang2025prompthmr}
& 71.0 & 216.5 & \rgbrankU{1.3}{100}
& 16.3 & \rgbrankU{3.5}{100}
& -- & -- & -- & -- & -- \\
GENMO~\cite{li2025genmo}
& \rgbrankU{65.1}{100}
& 210.9 & \rgbrankB{1.0}{100} & 9.6 & 8.3
& 80.7 & 127.2 & 2.6 & 8.6
& \rgbrankB{2.7}{100} \\
\midrule
TRAM + HTD-Refine~\cite{wang2024tram,wei2026natural}
& 71.7 & 204.9 & 1.5
& \rgbrankB{6.6}{100} & 7.5
& 90.2 & 145.3 & 2.5 & 4.2 & 6.5 \\
GVHMR + HTD-Refine~\cite{shen2024world,wei2026natural}
& 69.2 & 192.4 & 1.5 & 7.2 & 5.7
& 75.2 & 123.8 & \rgbrankU{2.3}{100}
& \rgbrankU{3.6}{100} & 3.3 \\
\midrule
\textbf{Ours w/ masked DROID-SLAM}
& 66.6 & \rgbrankU{189.8}{100}
& 1.4 & \rgbrankU{6.7}{100} & 5.4
& -- & -- & -- & -- & -- \\
\textbf{Ours w/ ViPE}
& \rgbrankB{57.4}{100}
& \rgbrankB{154.0}{100}
& \rgbrankU{1.3}{100}
& 8.4 & 5.2
& \rgbrankB{65.5}{100}
& \rgbrankB{109.1}{100}
& \rgbrankB{2.1}{100}
& \rgbrankB{3.5}{100}
& 4.3 \\
\bottomrule
\end{tabular}%
}
\end{table*}

\begin{figure*}[t]
    \centering
    \includegraphics[width=0.93\textwidth]{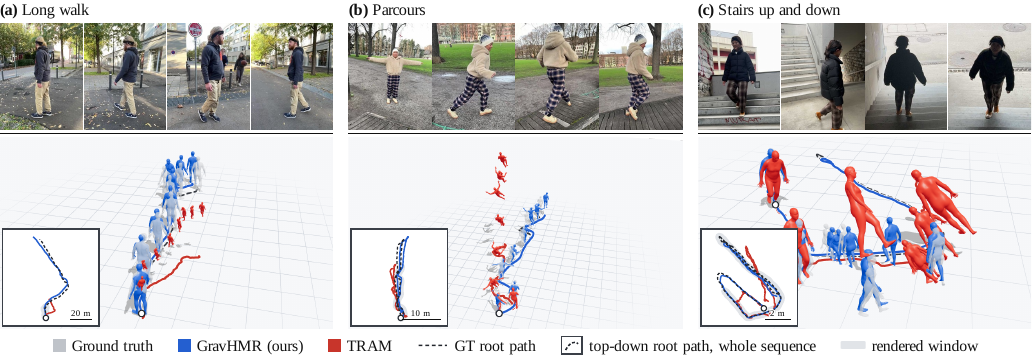}
    \caption{\textbf{Qualitative comparison of global human motion.}
    GT (\colorswatch{gtgray}), GravHMR (\colorswatch{oursblue}), and TRAM (\colorswatch{tramred})
    on three EMDB-2 sequences.
    GravHMR uses ViPE cameras; TRAM uses its masked DROID-SLAM cameras
    with ground-truth intrinsics. Each rendered 15\,s window is aligned
    to the ground truth on its first 60 frames.
    Insets: top-down root paths of the whole sequence.}
    \label{fig:global_motion_qual}
\end{figure*}

\begin{figure}
  \centering
  \includegraphics[width=\linewidth]{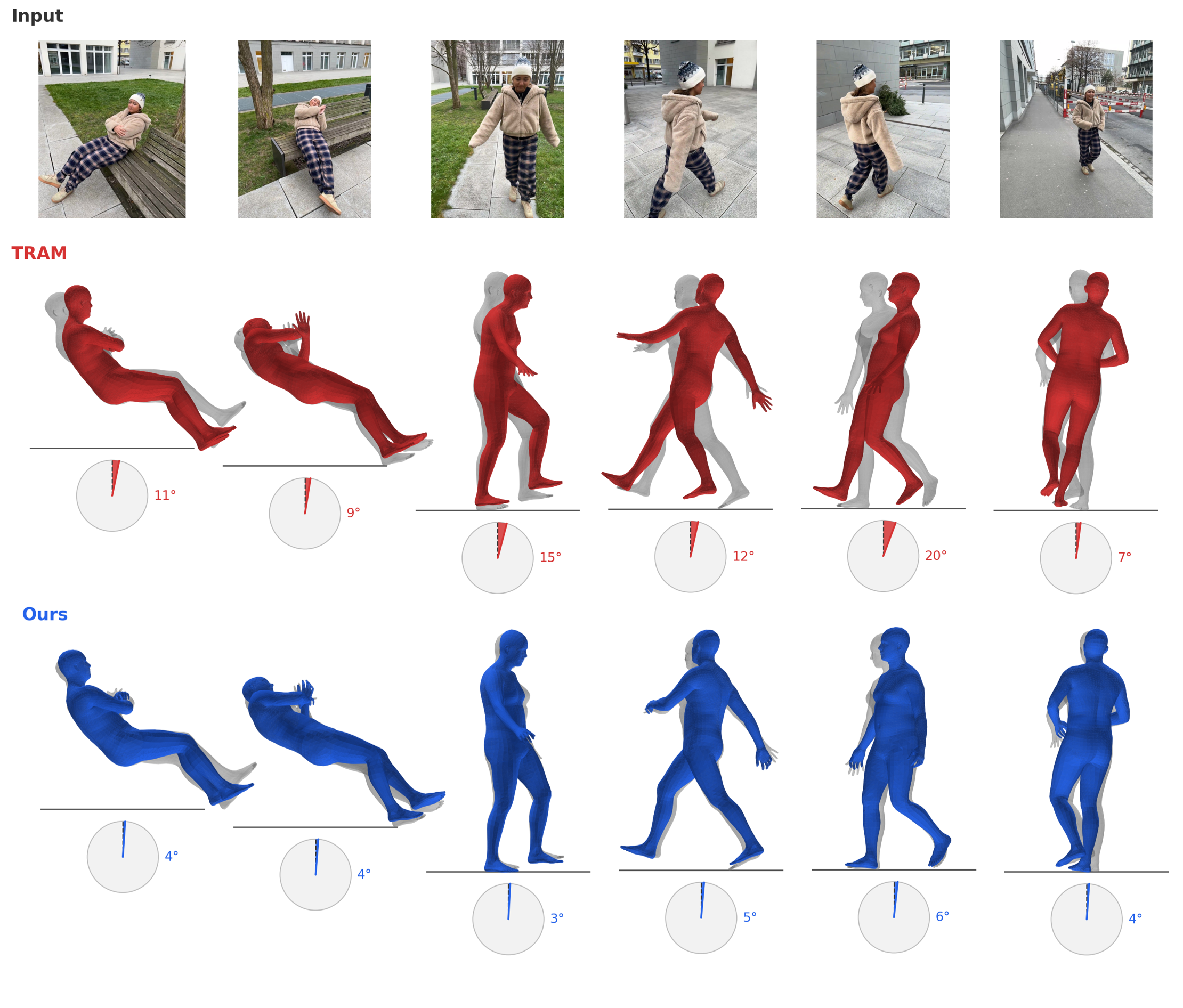}
  \caption{\textbf{Gravity-aligned coupling improves body pose and root orientation.}
  Qualitative comparison on EMDB-2 sequence \texttt{P7\_61}.
  TRAM (top, red) and ours (bottom, blue) are overlaid on the
  ground truth (gray), translated to the ground-truth pelvis
  without rotational alignment.}
  \label{fig:root_orientation}
\end{figure}

\begin{table}[t]
\centering
\caption{\textbf{Ablation of the core reconstruction on EMDB-2.}
All variants use the same ViPE camera poses and intrinsics and are
evaluated before FATC.
``w/o YAOO body'' keeps the rollout body orientation but the
yaw-corrected motion prior; ``w/o YAOO body \& motion'' reverts both
to the rollout. Each GATO ablation drops one loss term and
re-optimizes. The last row adds FATC to the full core and is not ranked.
Best variant in bold, except for Jitter (lower may indicate
over-smoothing).}
\label{tab:core_ablation}
\setlength{\tabcolsep}{4pt}
\renewcommand{\arraystretch}{1.10}
\resizebox{\columnwidth}{!}{%
\begin{tabular}{lccccc}
\toprule
Method
& WA$_{100}\downarrow$
& W$_{100}\downarrow$
& ERVE  $\downarrow$
& Jitter
& FS    $\downarrow$ \\
\midrule
\rowcolor{blockrow}
\multicolumn{6}{l}{\textit{Reference configurations}} \\
GVHMR rollout
& 106.7 & 262.8 & 11.6 & 15.2 & 3.4 \\
\midrule
\rowcolor{blockrow}
\multicolumn{6}{l}{\textit{YAOO ablations}} \\
w/o YAOO body
& 62.7 & 176.9 & 10.8 & 35.0 & 6.3 \\
w/o YAOO body \& motion
& 64.1 & 191.5 & 11.1 & 36.1 & 6.5 \\
\midrule
\rowcolor{blockrow}
\multicolumn{6}{l}{\textit{GATO ablations}} \\
w/o $\mathcal{L}_{\text{reproj}}$
& 88.1 & 202.3 & 13.1 & 46.4 & 8.4 \\
w/o $\mathcal{L}_{\text{human-scale}}$
& 59.0 & 165.3 & 10.6 & 30.7 & 7.5 \\
w/o $\mathcal{L}_{\text{dynamic}}$
& 57.7 & 165.7 & 10.9 & 41.6 & 9.5 \\
\midrule
Full core
& \textbf{57.6} & \textbf{155.3} & \textbf{10.2} & 32.1 & 7.8 \\
\quad + FATC
& 57.4 & 154.0 & 9.2 & 8.4 & 5.2 \\
\bottomrule
\end{tabular}%
}
\end{table}

\begin{table}[t]
\centering
\small
\caption{\textbf{System-level ablation of GravHMR.}
Each pair uses identical backbone predictions and ViPE camera
parameters, evaluated before FATC.
$E_{\text{tilt}}$, $E_{\text{yaw}}$, and $E_{\text{dir}}$
measure body tilt, heading, and travel-direction errors in degrees.}
\label{tab:system_ablation}
\setlength{\tabcolsep}{4pt}
\renewcommand{\arraystretch}{1.08}
\resizebox{\columnwidth}{!}{%
\begin{tabular}{lccc ccc}
\toprule
\multirow{2}{*}{Method}
& \multicolumn{3}{c}{Global motion}
& \multicolumn{3}{c}{Orientation (deg)} \\
\cmidrule(lr){2-4}\cmidrule(lr){5-7}
& WA$_{100}\downarrow$ & W$_{100}\downarrow$ & RTE$\downarrow$
& $E_{\text{tilt}}$ & $E_{\text{yaw}}\downarrow$ & $E_{\text{dir}}\downarrow$ \\
\midrule
\rowcolor{blockrow}
\multicolumn{7}{l}{\textit{EMDB-2}} \\
w/o GravHMR
& 106.7 & 262.8 & 2.10 & 8.14 & 7.33 & 11.8 \\
w/ GravHMR
& \textbf{57.6} & \textbf{155.3} & \textbf{1.27}
& 8.18 & \textbf{6.69} & \textbf{5.4} \\
\midrule
\rowcolor{blockrow}
\multicolumn{7}{l}{\textit{EgoBody}} \\
w/o GravHMR
& 129.4 & 203.5 & 3.85 & 5.73 & 7.04 & 30.2 \\
w/ GravHMR
& \textbf{87.5} & \textbf{123.3} & \textbf{1.04}
& 5.21 & \textbf{6.76} & \textbf{9.3}\\
\midrule
\rowcolor{blockrow}
\multicolumn{7}{l}{\textit{EMDB-2, VIMO backbone (no gravity view)}} \\
w/o GravHMR
& 73.3 & 197.9 & 1.28 & 8.18 & 6.55 & 5.7 \\
w/ GravHMR
& \textbf{71.5} & \textbf{196.3} & 1.28 & 8.18 & 6.55 & 5.7 \\
\bottomrule
\end{tabular}%
}
\end{table}

\begin{figure*}[t] \centering \includegraphics[width=0.95\textwidth]{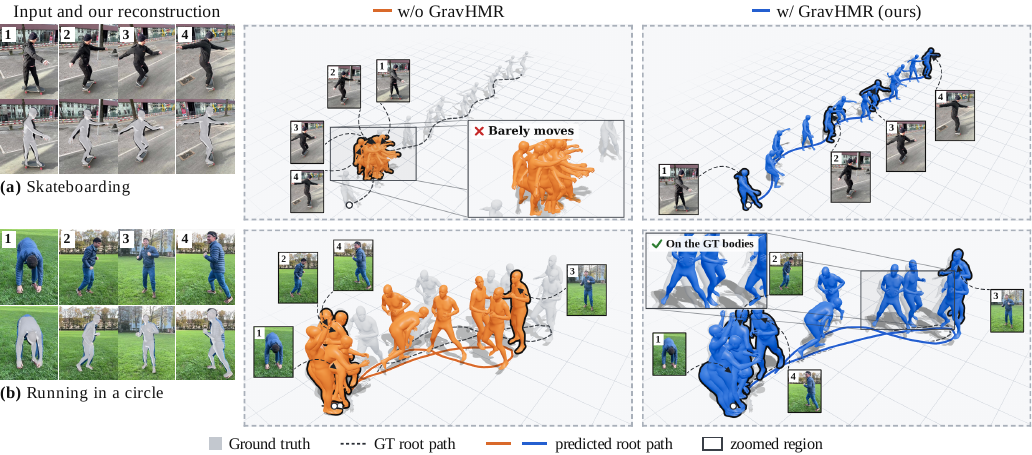} \caption{ \textbf{GravHMR corrects global motion beyond locally plausible poses.} We compare reconstructions without (orange) and with (blue) GravHMR on two EMDB-2 sequences. Gray meshes and dashed curves indicate ground-truth bodies and root trajectories, respectively.} \label{fig:supp_wwo_gravhmr} \end{figure*}

\subsection{Experimental Setup}
\label{sec:setup}

\noindent\textbf{Datasets.}
We evaluate world-grounded motion on EMDB-2~\cite{kaufmann2023emdb},
25 in-the-wild sequences captured by a moving hand-held camera, and on
the RICH~\cite{huang2022capturing} test set, 191 views recorded by
static cameras. For the ablations we additionally use
EgoBody~\cite{zhang2022egobody} (17 recordings), whose head-mounted
camera frequently truncates the body.

\noindent\textbf{Metrics.}
Following GVHMR~\cite{shen2024world}, we report
W-MPJPE$_{100}$ and WA-MPJPE$_{100}$ for global
joint accuracy over $100$-frame segments, using
alignment based on the first two frames and the
entire segment, respectively.
RTE measures trajectory error normalized by
ground-truth travel distance.
For camera-space reconstruction, we report PA-MPJPE, MPJPE, and PVE.
Accel, Jitter, and Foot-Sliding (FS) measure motion smoothness and foot
stability, and ERVE and W-RVE evaluate root-velocity accuracy.
Lower is better for all metrics.

\subsection{World-Grounded Performance}
\label{sec:exp_world}

\noindent\textbf{Trajectory Accuracy.}
GravHMR attains the lowest WA- and W-MPJPE$_{100}$ on both datasets
(Tab.~\ref{tab:world}): $57.4$ and $154.0$\,mm on EMDB-2, and $65.5$
and $109.1$\,mm on RICH.
With the same HMR backbone, it reduces W-MPJPE$_{100}$ relative to
GVHMR + HTD-Refine by $38.4$\,mm ($20.0\%$) on EMDB-2 and $14.7$\,mm
($11.9\%$) on RICH; on RICH it also improves on the strongest prior
result, GVHMR + PVA-Net, by $10.0$\,mm ($8.4\%$).
RTE reaches $1.3\%$ and $2.1\%$; only GENMO attains a lower RTE on
EMDB-2 ($1.0\%$).

\noindent\textbf{Effect of FATC.}
FATC acts on the core reconstruction (last row of
Tab.~\ref{tab:core_ablation}) and its effect is mainly temporal:
on EMDB-2 it reduces Jitter from $32.1$ to $8.4$ and FS from $7.8$ to
$5.2$\,mm while W-MPJPE$_{100}$ moves only from $155.3$ to
$154.0$\,mm. On RICH it reduces Jitter from $33.8$ to $3.5$ and FS
from $6.06$ to $4.32$\,mm, and also lowers W-MPJPE$_{100}$ from
$119.0$ to $109.1$\,mm.

\subsection{Qualitative Comparison}
\label{sec:qualitative}

\noindent\textbf{Body Tilt and Vertical Alignment.}
Panel~(c) of Fig.~\ref{fig:global_motion_qual} shows pronounced body
tilt and vertical misalignment in the TRAM reconstruction, whereas
GravHMR closely follows the ground-truth body orientation and height
from both viewpoints.
Fig.~\ref{fig:root_orientation} isolates root-relative errors on
EMDB-2 sequence P7\_61. TRAM deviates from the ground-truth body axis
by up to $24^\circ$ in the displayed frames, against $3$--$8^\circ$
for GravHMR; over the full sequence, the median body-axis error falls
from $12.4^\circ$ to $4.9^\circ$.
TRAM lifts the camera-frame body orientation with the SLAM rotation
and thus inherits the tilt error of the camera, whereas YAOO keeps
the gravity-relative tilt of the rollout (Sec.~\ref{sec:yaoo}).

\noindent\textbf{Travel Direction and Turns.}
Panels~(a) and~(b) of Fig.~\ref{fig:global_motion_qual} show the long
walk P2\_24 and the parcours P7\_58, in which the subject repeatedly
changes heading. TRAM poses the body plausibly frame by frame, but its
path departs from the reference, and the top-down insets show turns
rounded off or taken at the wrong heading.
GravHMR anchors the root to the 2D keypoints in every frame under the
metric SLAM camera instead of integrating a predicted velocity, so
heading errors do not accumulate and the turns stay on the reference.
Relative to the GVHMR rollout, the direction-of-travel error
$E_{\text{dir}}$ falls from $11.8^\circ$ to $5.4^\circ$
(Tab.~\ref{tab:system_ablation}).

\noindent\textbf{Global Trajectory of the HMR Backbone.}
This comparison isolates what GravHMR adds to its own backbone.
Fig.~\ref{fig:supp_wwo_gravhmr} and the whole-sequence top-down
trajectories of Fig.~\ref{fig:traj_topdown} compare the GVHMR rollout
before and after GravHMR, each aligned to the ground truth by a single
Sim(3) over the first $100$ frames. The rollout fails in two ways.
On the rectangular walk and the stairs, it keeps the shape of the
path but accumulates a steady offset, the signature of a trajectory
integrated from predicted velocities.
On the skateboard sequence, the subject covers a long distance while
the body is nearly static in the image, so the motion prior has no
evidence of travel and the root stays where it started.
GravHMR reads the root from the camera and the image instead of
integrating it, reducing the mean horizontal error to $0.2$--$0.3$\,m
on the walking and stair sequences and from $17.3$ to $1.3$\,m on the
skateboard sequence.
Fig.~\ref{fig:supp_wild} in the appendix shows in-the-wild running,
skateboarding, skating, cycling, and basketball videos, where no
ground truth is available.

\subsection{Comparison under Identical Camera Inputs}
\label{sec:shared_camera}

\noindent\textbf{Does better SLAM yield better global HMR performance?}
Improved camera estimates benefit global human motion
recovery, but the reconstruction pipeline also matters.
Using the same masked DROID-SLAM camera estimates
and ground-truth intrinsics as TRAM,
GravHMR achieves 189.8\,mm W-MPJPE$_{100}$
on EMDB-2, compared with 222.4\,mm for TRAM
and 204.9\,mm for TRAM + HTD-Refine,
reducing error by 14.7\% and 7.4\%, respectively.
With ViPE camera poses and estimated intrinsics,
GravHMR further reduces the error to 154.0\,mm.
These results demonstrate gains under identical
camera inputs and additional benefits from
the ViPE camera configuration.


\begin{table}[t]
\centering
\small
\caption{\textbf{How much of the camera-derived orientation to adopt.}
\emph{Rollout} keeps $\mathbf{R}^{\mathrm{roll}}_t$ for the body and
the motion prior; \emph{yaw only} is YAOO
(Eq.~\eqref{eq:yaoo_apply}); \emph{full} uses $\widetilde{\mathbf{R}}_t$
of Eq.~\eqref{eq:yaoo_ref} as the body orientation.
All rows share the ViPE camera, the HMR prediction and GATO, and are
evaluated before FATC on EMDB-2.
$E_{\text{dir}}$ is identical across the three rows.}
\label{tab:orient_src}
\setlength{\tabcolsep}{5pt}
\renewcommand{\arraystretch}{1.08}
\resizebox{\columnwidth}{!}{%
\begin{tabular}{l c ccc}
\toprule
\multirow{2}{*}{Body orientation $\mathbf{R}_t$}
& \multirow{2}{*}{W$_{100}\downarrow$}
& \multicolumn{3}{c}{Orientation (deg)} \\
\cmidrule(lr){3-5}
& & $E_{\text{tilt}}\downarrow$ & $E_{\text{yaw}}\downarrow$ & $E_{\text{dir}}\downarrow$ \\
\midrule
Rollout (w/o YAOO)
& 191.5 & 8.34 & 7.31 & 5.4 \\
Full camera orientation
& 165.7 & 8.34 & 6.76 & 5.4 \\
Yaw only (YAOO, ours)
& \textbf{155.3} & \textbf{8.18} & \textbf{6.69} & 5.4 \\
\bottomrule
\end{tabular}%
}
\end{table}

\begin{table}[t]
\centering
\small
\caption{\textbf{Source of the root depth along the viewing ray}
(Eq.~\eqref{eq:init} and Eq.~\eqref{eq:Lhs}), on EMDB-2 and EgoBody
with ViPE cameras, before FATC.
\emph{2D body extent} takes the depth of the per-frame 2D root solve;
\emph{HMR depth} takes $z^{\mathrm{HMR}}_t$.
The first row is the GVHMR rollout and uses neither.}
\label{tab:depth_src}
\setlength{\tabcolsep}{5pt}
\renewcommand{\arraystretch}{1.08}
\resizebox{\columnwidth}{!}{%
\begin{tabular}{l cc cc}
\toprule
\multirow{2}{*}{Depth source}
& \multicolumn{2}{c}{EMDB-2}
& \multicolumn{2}{c}{EgoBody} \\
\cmidrule(lr){2-3}\cmidrule(lr){4-5}
& WA$_{100}\downarrow$ & W$_{100}\downarrow$
& WA$_{100}\downarrow$ & W$_{100}\downarrow$ \\
\midrule
w/o GravHMR
& 106.7 & 262.8 & 129.4 & 203.5 \\
\midrule
2D body extent
& 57.7 & 155.3 & 133.0 & 205.0 \\
HMR depth
& \textbf{57.6} & \textbf{155.3} & \textbf{87.5} & \textbf{123.3} \\
\bottomrule
\end{tabular}%
}
\end{table}

\subsection{Ablation Study}
\label{sec:ablation}

\noindent\textbf{YAOO and GATO.}
Table~\ref{tab:core_ablation} evaluates the core
components of GravHMR on EMDB-2.
All variants use the same ViPE camera estimates,
body pose, and shape, and are evaluated before FATC.
Each ablation removes or replaces one component of the
full core while retaining the remaining settings.

\noindent\textbf{Role of YAOO.}
Tab.~\ref{tab:orient_src} asks how much of the camera-derived
orientation should be adopted.
Keeping the rollout orientation leaves its heading drift in both the
body and the motion prior: W-MPJPE$_{100}$ rises from $155.3$ to
$191.5$\,mm and $E_{\text{yaw}}$ from $6.69^\circ$ to $7.31^\circ$.
Adopting $\widetilde{\mathbf{R}}_t$ in full corrects the heading
($E_{\text{yaw}}$ $6.76^\circ$) but also imports its tilt
($E_{\text{tilt}}$ $8.34^\circ$ vs.\ $8.18^\circ$), and
W-MPJPE$_{100}$ rises to $165.7$\,mm.
Transferring only the yaw keeps the camera heading and the
gravity-view tilt, and gives the lowest error.
Tab.~\ref{tab:core_ablation} further separates where YAOO acts:
reverting only the body orientation gives $176.9$\,mm, and reverting
the motion prior as well gives $191.5$\,mm, so the corrected heading
matters both for the reprojected joints and for the direction of the
motion prior.
$E_{\text{dir}}$ is $5.4^\circ$ in all rows of Tab.~\ref{tab:orient_src}:
the direction of travel is fixed by GATO, not by YAOO.

\noindent\textbf{Role of GATO.}
The trajectory constraints serve different purposes.
Removing the reprojection loss increases
W-MPJPE$_{100}$ from $155.3$ to $202.3$\,mm,
showing its importance for maintaining geometric
accuracy during trajectory optimization.
The motion prior improves both global accuracy and
temporal consistency: removing it raises W-MPJPE$_{100}$
from $155.3$ to $165.7$\,mm and increases ERVE
from $10.2$ to $10.9$, Jitter from $32.1$ to $41.6$,
and FS from $7.8$ to $9.5$\,mm.
Replacing the gravity-view displacement prior
with camera-frame displacements transformed
into the world frame further increases ERVE
to $11.3$, Jitter to $46.9$, and FS to $10.0$\,mm,
supporting the use of gravity-referenced motion
predictions for motion consistency.
Removing the human-scale depth prior raises
W-MPJPE$_{100}$ from $155.3$ to $165.3$\,mm and
WA-MPJPE$_{100}$ from $57.6$ to $59.0$\,mm,
while slightly reducing Jitter and FS.

\noindent\textbf{Source of the Depth Prior in GATO.}
Tab.~\ref{tab:depth_src} replaces the HMR depth of Eq.~\eqref{eq:init}
and Eq.~\eqref{eq:Lhs} by the depth of the per-frame 2D root solve,
\ie the body extent in the image. On EMDB-2 the two are equivalent
($155.3$\,mm both), since the subject is fully visible and the body
extent is a reliable ruler. On EgoBody the 2D solve gives $205.0$\,mm,
close to the $203.5$\,mm of the rollout it starts from, while the HMR
depth reaches $123.3$\,mm: under the frequent truncation of a
head-mounted camera the body extent is no longer measurable in the
image, whereas HMR still reports a depth.

\noindent\textbf{Sensitivity of the Reprojection Term in GATO.}
Replacing the self-calibrated scale with a constant ($13.8$\,px),
removing the confidence weights, or replacing the Cauchy penalty
with squared residuals yields $155.3$, $155.3$, and $156.0$\,mm
W-MPJPE$_{100}$, respectively, compared with $155.3$\,mm for
the full core. These changes have little effect on global joint
accuracy on EMDB-2, whereas removing the reprojection term
entirely increases the error to $202.3$\,mm
(Tab.~\ref{tab:core_ablation}). This suggests that retaining
the image-based geometric constraint matters more in this setting
than the particular scaling, weighting, or penalty tested here.

\noindent\textbf{Coupling of Orientation and Trajectory Recovery.}
YAOO and GATO interact: the body orientation determines the
root-relative joints used for reprojection and the world-space
direction of the motion prior, and GATO recovers the root under this
orientation. Together they reduce W-MPJPE$_{100}$ from $262.8$ to
$155.3$\,mm relative to the GVHMR rollout, before FATC
(Tab.~\ref{tab:system_ablation}), and from $203.5$ to $123.3$\,mm
($-39.4\%$) on EgoBody.
The orientation columns separate their roles: YAOO only rotates about
gravity, so $E_{\text{tilt}}$ stays within $0.05^\circ$ of the rollout
on EMDB-2 while $E_{\text{yaw}}$ falls from $7.33^\circ$ to
$6.69^\circ$; $E_{\text{dir}}$, governed by GATO, falls from
$11.8^\circ$ to $5.4^\circ$. With VIMO~\cite{wang2024tram}, which has no gravity-view head,
GravHMR changes W-MPJPE$_{100}$ only from $197.9$ to $196.3$\,mm
(Tab.~\ref{tab:system_ablation}). VIMO's world-space body is already
the camera-frame orientation lifted by the SLAM rotation and its root
is already camera-anchored, so YAOO has no rollout heading to correct
and GATO little drift to remove. The large gains on GVHMR therefore
come from reconciling a gravity-view rollout with the camera, which is
the setting GravHMR targets.

\section{Conclusion}
\label{sec:conclusion}

We presented GravHMR, a training-free framework
that couples SLAM camera geometry and HMR motion
priors through a shared gravity-aligned reference.
Our approach selectively corrects body and velocity
headings while retaining gravity-referenced body tilt.
It further uses 2D pose observations to refine pose
and orientation before recovering global root trajectories,
reducing inconsistencies between human motion priors
and camera geometry without retraining the HMR backbone.

On EMDB-2, GravHMR reaches 154.0\,mm W-MPJPE$_{100}$,
reducing error by 44.3\% relative to GVHMR and 20.0\%
relative to GVHMR with HTD-Refine, and 189.8\,mm under
the masked DROID-SLAM cameras used by TRAM, an improvement
of 14.7\% over it.
On RICH it reaches 109.1\,mm, improving upon GVHMR with
HTD-Refine by 11.9\%.
These results support gravity-constrained coupling
and staged orientation--trajectory recovery as an
effective approach to world-grounded human motion reconstruction.

\section*{LLM Usage Statement}

OpenAI ChatGPT was used to assist with language editing,
LaTeX formatting, and manuscript organization.
Anthropic Claude was used to assist with experimental
validation: running evaluation scripts, cross-checking the
reported numbers against the run logs.
The authors reviewed and verified the final manuscript
and remain responsible for all scientific claims,
experimental results, and conclusions.

{
    \small

}

\clearpage
\appendix
\setcounter{dbltopnumber}{4}
\renewcommand{\dbltopfraction}{0.95}
\renewcommand{\dblfloatpagefraction}{0.8}

\renewcommand{\thefigure}{A\arabic{figure}}
\renewcommand{\thetable}{A\arabic{table}}
\renewcommand{\theequation}{A\arabic{equation}}
\setcounter{equation}{0}
\setcounter{figure}{0}
\setcounter{table}{0}
\setcounter{section}{0}

\maketitlesupplementary

\noindent
This appendix provides additional analyses and implementation
details that complement the main paper.
Appendix~\ref{appendix:impl} gives implementation details:
the notation (Appendix~\ref{appendix:notation}),
how YAOO and GATO share one heading (Appendix~\ref{appendix:angle})
and the per-joint weighting and scales of GATO
(Appendix~\ref{appendix:jointweight}).
Appendix~\ref{appendix:nongv} discusses HMR backbones without a
gravity view, and Appendix~\ref{appendix:results} reports the
remaining results.
Fig.~\ref{fig:supp_wild} compares the HMR backbone with and
without GravHMR on in-the-wild videos.

\section{Implementation Details}
\label{appendix:impl}

\subsection{Notation}
\label{appendix:notation}

The symbols used in this appendix follow the main paper.
\begin{center}
\small
\setlength{\tabcolsep}{4pt}
\renewcommand{\arraystretch}{1.12}
\begin{tabular}{@{}l p{0.70\columnwidth}@{}}
\toprule
Symbol & Meaning \\
\midrule
$\mathbf{K}_t,\ \mathbf{R}^{c\to g}_t,\ \mathbf{C}^{g}_t$ & SLAM intrinsics, camera rotation and camera center in the gravity frame (Sec.~\ref{sec:gv_slam_hmr}) \\
$\mathbf{n}^{g}$ & gravity axis of the gravity frame, $[0,1,0]^\top$ (Eq.~\eqref{eq:gravity_axis}) \\
$\boldsymbol{\Gamma}^c_t$ & HMR camera-frame root orientation \\
$\mathbf{R}^{\mathrm{roll}}_t,\ \mathbf{X}^{\mathrm{roll}}_t$ & orientation and root position of the HMR rollout \\
$\widetilde{\mathbf{R}}_t$ & camera-frame orientation lifted into the gravity frame (Eq.~\eqref{eq:yaoo_ref}) \\
$\psi_t$ & yaw about $\mathbf{n}^{g}$ that aligns $\mathbf{R}^{\mathrm{roll}}_t$ with $\widetilde{\mathbf{R}}_t$ (Eq.~\eqref{eq:yaoo_psi}) \\
$\mathbf{R}_t,\ \mathbf{d}_t$ & body orientation and motion prior after YAOO (Eq.~\eqref{eq:yaoo_apply}) \\
$\mathbf{r}_t,\ z^{\mathrm{HMR}}_t$ & viewing direction of the per-frame 2D root solve and HMR camera-frame depth (Eq.~\eqref{eq:init}) \\
$\mathbf{X}_t$ & world-space root position recovered by GATO (Eq.~\eqref{eq:GATO}) \\
$\mathbf{u}_{t,k},\ c_{t,k}$ & 2D keypoint $k$ at frame $t$ and its detector confidence (Eq.~\eqref{eq:Lreproj}) \\
$\sigma_{\mathrm{2D}},\ \sigma_{z},\ \sigma_{d}$ & self-calibrated scales of the three GATO terms (Appendix~\ref{appendix:sigma}) \\
\bottomrule
\end{tabular}
\end{center}

\subsection{Angular Consistency Between YAOO and GATO}
\label{appendix:angle}

GATO (Sec.~\ref{sec:GATO}) uses the body orientation in two places:
the reprojection term of Eq.~\eqref{eq:Lreproj} places the
root-relative joints with $\mathbf{R}_t$, and the dynamic term of
Eq.~\eqref{eq:Ldyn} compares the root displacement with
$\mathbf{d}_t$. Both come from Eq.~\eqref{eq:yaoo_apply} and are
rotated by the same yaw $\psi_t$, so the two terms see one heading
and cannot pull the root toward different directions. The depth term
of Eq.~\eqref{eq:Lhs} does not depend on the heading. As stated in
Sec.~\ref{sec:GATO}, the three terms are summed without weights,
each with its own self-calibrated scale (Appendix~\ref{appendix:sigma}).
Tab.~\ref{tab:core_ablation} shows what happens when the shared
heading is broken: reverting only the body orientation to the rollout
gives $176.9$\,mm W-MPJPE$_{100}$, and reverting the motion prior as
well gives $191.5$\,mm, against $155.3$\,mm for the full core.

\subsection{Per-Joint Weighting in GATO}
\label{appendix:jointweight}

This section details the per-joint weighting strategies compared
in Tab.~\ref{tab:joint_weight}. All variants share the same
GATO pipeline and differ only in the per-joint weights
$w_{t,k}$ used in the reprojection term; the main results
use the detector confidence, $w_{t,k}=c_{t,k}$, as in
Eq.~\eqref{eq:Lreproj}.
GATO uses the $K=17$ COCO keypoints
$\{\mathbf{u}_{t,k}\}_{k=1}^{K}$.
With the body orientation $\mathbf{R}_t$ fixed by YAOO,
the camera-frame root $\mathbf{c}_t$ of each frame is first
obtained in closed form. Writing
$\mathbf{b}_{t,k}=\mathbf{R}^{g\to c}_t\mathbf{R}_t\mathbf{J}_k$
for the root-relative joint position in the camera frame and
$(x_{t,k},y_{t,k})$ for the normalized image coordinates of
$\mathbf{u}_{t,k}$, each joint contributes two equations that
are linear in $\mathbf{c}_t$,
\begin{align}
(c_x + b_x) - x_{t,k}\,(c_z + b_z) &= 0, \nonumber\\
(c_y + b_y) - y_{t,k}\,(c_z + b_z) &= 0,
\label{eq:supp_lin}
\end{align}
and $\mathbf{c}_t$ is the weighted linear least-squares
solution of the $34$ equations, each scaled by $\sqrt{w_{t,k}}$.
This per-frame solve provides the viewing direction
$\mathbf{r}_t$ used to initialize the root and the reference
residuals used to self-calibrate $\sigma_{\mathrm{2D}}$.
The same weights, normalized to unit mean, are then frozen
and reused in the reprojection loss $\mathcal{L}_{\text{reproj}}$
(Eq.~\eqref{eq:Lreproj}) in place of $c_{t,k}$.
Given a root estimate $\mathbf{c}_t$, the reprojection
residual of joint $k$ at frame $t$ is
$r_{t,k}=\big\|\pi\big(\mathbf{K}_t(\mathbf{b}_{t,k}+\mathbf{c}_t)\big)
-\mathbf{u}_{t,k}\big\|_2$ in pixels.
Each joint's characteristic error is summarized by a robust
scale over the sequence,
\begin{equation}
\sigma_k \;=\; \max\!\Big(1.4826\cdot\!\!
\operatorname*{median}_{t\in\mathcal{V}_k} r_{t,k},\;
1\,\mathrm{px}\Big),
\label{eq:supp_sigk}
\end{equation}
where $\mathcal{V}_k$ is the set of frames in which joint $k$
is considered valid.
$\sigma_k$ absorbs both 2D detection noise and the error of
the root-relative body offset at the actual depth, and is
estimated from the data without ground truth.

\paragraph{Scales of the GATO terms.}
\label{appendix:sigma}
The scale of each term of Eq.~\eqref{eq:GATO} is measured once on
the input, before optimization. Let $\mathbf{X}^{\text{2D}}_t$
and $z^{\text{2D}}_t$ be the root and its depth from the per-frame 2D root solve.
$\sigma_{\mathrm{2D}}$ (Eq.~\eqref{eq:Lreproj}) is the median absolute
deviation (MAD) of the reprojection residuals of that solve;
$\sigma_{z}$ (Eq.~\eqref{eq:Lhs}) is set per frame to
$z^{\text{2D}}_t\,\mathrm{MAD}\big(\log(z^{\text{2D}}_t/z^{\mathrm{HMR}}_t)\big)$,
which compares the depth of that solve with the HMR depth; and
$\sigma_{d}$ (Eq.~\eqref{eq:Ldyn}) is the MAD of
$\|(\mathbf{X}^{\text{2D}}_{t+1}-\mathbf{X}^{\text{2D}}_t)-\mathbf{d}_t\|$,
the disagreement between its displacements and the motion prior.

\paragraph{Structural and dynamic weights.}
The structural weight $\pi_k$ favors joints that lie close to
the root and are rarely truncated: $\pi_k=1$ for the
shoulders and hips (COCO indices 5, 6, 11, 12) and
$\pi_k=\lambda=0.25$ for all other joints. Used alone
(``Structural'' in Tab.~\ref{tab:joint_weight}),
$w_{t,k}=\pi_k$.
The dynamic weight instead down-weights a joint only in the frames
where it disagrees with the body-coherent root direction, for
example when it is truncated, occluded, or mis-detected:
\begin{equation}
\alpha_{t,k} \;=\; \pi_k \cdot
\frac{\sigma_k^{2}}{\sigma_k^{2}+r_{t,k}^{2}},
\label{eq:supp_dyn}
\end{equation}
with $\pi_k\equiv 1$ for ``Dynamic'' alone and $\pi_k$ as
above for ``Structural + Dynamic''.
Eq.~\eqref{eq:supp_dyn} is the IRLS weight of a Cauchy loss
with a per-joint scale $\sigma_k$, so each joint is judged
against its own typical error rather than against a scale
shared by all joints.
The dynamic weights depend on the root through $r_{t,k}$,
so they are estimated by iteratively reweighted least squares.
We start from the uniform (or, with the prior, structural)
solve, and repeatedly
(i) compute $r_{t,k}$,
(ii) update $\sigma_k$ by Eq.~\eqref{eq:supp_sigk},
(iii) update $\alpha_{t,k}$ by Eq.~\eqref{eq:supp_dyn},
(iv) normalize the weights to unit mean over the sequence,
and (v) re-solve Eq.~\eqref{eq:supp_lin}.
Three iterations suffice for the weights to stabilize; the
procedure adds a negligible cost relative to the trajectory
optimization.
When fewer than three joints are valid in a frame, the solve
of Eq.~\eqref{eq:supp_lin} becomes ill-posed;
we then fall back to all joints with a small floor weight
($10^{-3}$) so that the frame keeps a defined root, which is
subsequently constrained by $\mathcal{L}_{\text{human-scale}}$
and $\mathcal{L}_{\text{dynamic}}$.

\begin{figure*}[!tp]
  \centering
  \includegraphics[width=0.7\textwidth]{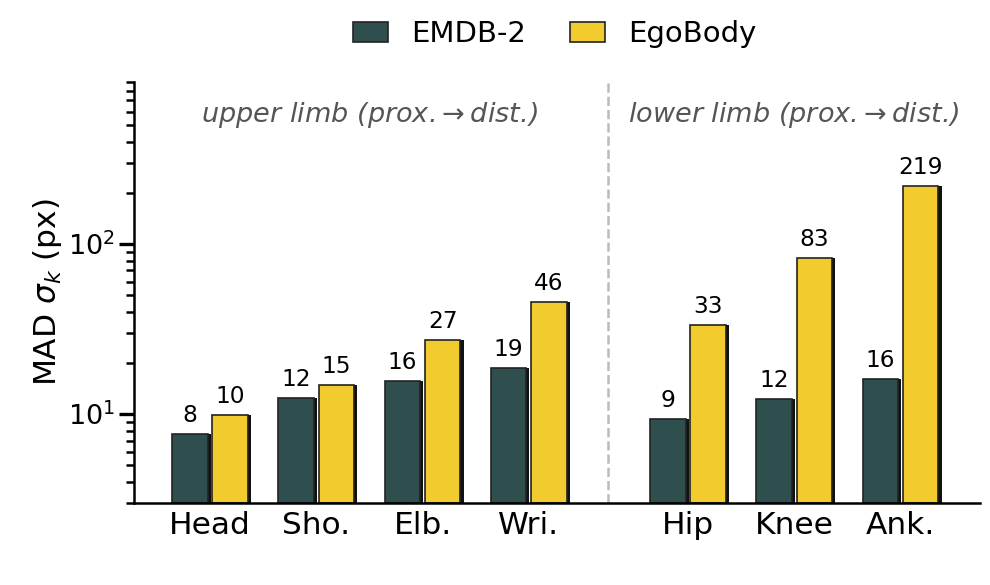}
  \vspace{-1em}
  \caption{\textbf{Per-joint reprojection reliability result.}
  MAD $\sigma_k$ of each joint's reprojection residual
  (median over sequences, log scale).
  Reliability degrades from proximal to distal joints on both datasets,
  and the degradation is amplified under truncation (EgoBody), most
  severely along the lower limb
  (ankle: 16\,px $\rightarrow$ 219\,px).
  This scene-dependent pattern motivates per-frame, per-joint weights
  rather than a fixed anchor set.}
  \label{fig:joint_mad}
  \vspace{-1em}
\end{figure*}

\paragraph{Effect of truncation (Tab.~\ref{tab:joint_weight}).}
Tab.~\ref{tab:joint_weight} was run with the depth of the per-frame
2D root solve (\emph{2D body extent} in Tab.~\ref{tab:depth_src}),
which makes GATO sensitive to which joints are trusted; its
detector-confidence row equals that row of Tab.~\ref{tab:depth_src}.
On EMDB-2, where the subject is mostly fully visible, all
strategies lie within $1.4$\,mm W-MPJPE$_{100}$.
Under frequent truncation (EgoBody), uniform weights give
$232.3$\,mm and the detector confidence $205.0$\,mm; the dynamic
weight lowers this to $144.9$\,mm, and combining it with the
structural prior to $136.1$\,mm, consistent with the per-joint
degradation in Fig.~\ref{fig:joint_mad}. With the HMR depth of the
main configuration, GravHMR reaches $123.3$\,mm on EgoBody with the
detector confidence alone (Tab.~\ref{tab:depth_src}).

\section{Generalization to HMR Backbones}
\label{appendix:nongv}

GVHMR predicts the body orientation in a gravity-view frame and
integrates a world-space rollout from it. Other HMR backbones, such as
VIMO~\cite{wang2024tram}, predict only a camera-frame orientation and
translation.

\paragraph{Source of the gravity reference.}
GravHMR takes gravity from the images with
GeoCalib~\cite{veicht2024geocalib} and expresses the SLAM cameras in
the resulting gravity frame (Sec.~\ref{sec:gv_slam_hmr}), so this
step does not depend on the backbone.

\paragraph{Backbones without a gravity view.}
For such a backbone, the world-space body is the camera-frame
orientation lifted by the SLAM rotation, so
$\mathbf{R}^{\mathrm{roll}}_t=\widetilde{\mathbf{R}}_t$ in
Eq.~\eqref{eq:yaoo_psi}, $\psi_t=0$, and YAOO leaves the body and the
motion prior unchanged; the world-space root is likewise the
camera-frame root lifted by the SLAM camera. GATO is applied without
change. With VIMO, GravHMR changes W-MPJPE$_{100}$ only from
$197.9$ to $196.3$\,mm on EMDB-2 (Tab.~\ref{tab:system_ablation}):
there is no rollout heading to correct and little drift to remove.
The large gains on GVHMR come from reconciling a gravity-view rollout
with the camera, which is the setting GravHMR targets.

\section{Additional Results}
\label{appendix:results}

\paragraph{Camera-space accuracy.}
GravHMR corrects the world-space orientation and trajectory
and refines body pose only through FATC.
Tab.~\ref{tab:rich_cam} checks that this does not cost
camera-space accuracy: on RICH, GravHMR improves PA-MPJPE
from 39.5 to 35.5\,mm and Accel from 4.1 to 3.35\,m/s$^2$
over GVHMR.

\paragraph{Camera intrinsics.}
ViPE estimates the focal length more accurately than TRAM on all 25 EMDB-2
sequences: the mean absolute error is 0.6\% against 7.7\%, and the maximum is
1.2\% against 34.6\% (Tab.~\ref{tab:emdb2_intrinsic}).
At this accuracy the remaining error no longer matters: replacing the ViPE
intrinsics with ground truth changes W-MPJPE$_{100}$ from 154.0 to 154.1\,mm.
TRAM's intrinsics, in contrast, raise its W-MPJPE$_{100}$ from 222.2 to 304.9\,mm.
GravHMR needs good intrinsics as well: with masked DROID-SLAM and TRAM's
intrinsics, it degrades from 189.8 to 270.4\,mm.

\paragraph{Gravity source.}
Tab.~\ref{tab:gravity_oracle} replaces the vertical of the body orientation,
which YAOO keeps from the HMR rollout (Sec.~\ref{sec:yaoo}), with GeoCalib or
ground-truth gravity. GeoCalib matches the rollout vertical on
W-MPJPE$_{100}$ (both $155.3$\,mm) and ground-truth gravity is slightly
worse ($160.0$\,mm), so the rollout vertical is not the limiting factor on
EMDB-2.

\paragraph{Runtime.}
Tab.~\ref{tab:runtime} lists the runtime of each stage. YAOO and GATO take about
10\,s per sequence; the SLAM and feature stages dominate the total time.

\paragraph{Component swap between rollout and GravHMR.}
Tab.~\ref{tab:ablation_yaoo_GATO} attributes the final trajectory to
the two modules without re-optimizing. Each per-frame root displacement
is split with respect to gravity into its heading (the body and travel
yaw set by YAOO) and its horizontal speed and vertical component (set
by GATO); each part is taken from the GVHMR rollout or from the full
core, and the trajectory is re-integrated. This differs from
Tab.~\ref{tab:core_ablation}, where GATO is re-run after YAOO is
removed. Taking only one part from GravHMR recovers little of the gain
(EMDB-2: $247.4$ and $245.3$\,mm W-MPJPE$_{100}$), and both parts
together give $155.3$\,mm. On EgoBody, GATO's speed and vertical
motion with the rollout heading is worse than the rollout itself
($234.7$ vs.\ $203.5$\,mm), while both parts give $123.3$\,mm.

\begin{figure*}[!tp]
  \centering
  \includegraphics[width=\textwidth]{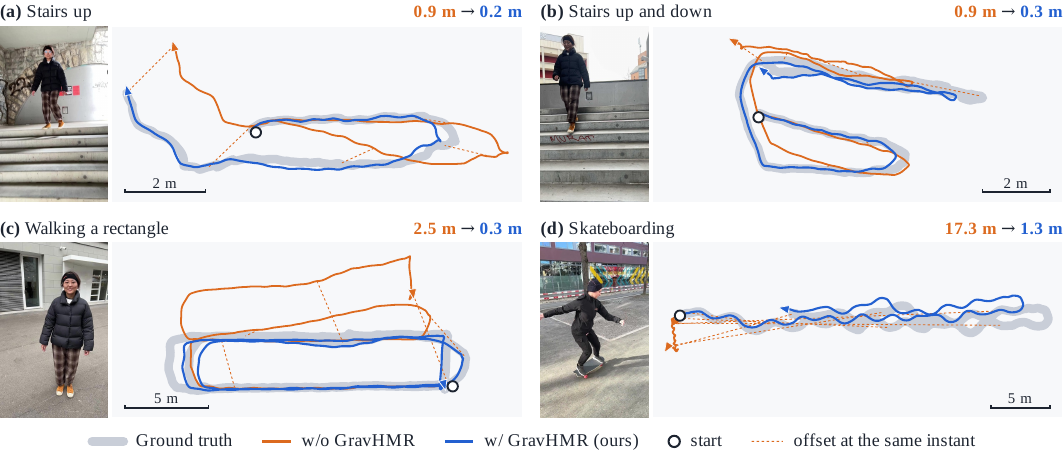}
  \caption{\textbf{Global trajectories of the HMR backbone with and without GravHMR.}
  Top-down root paths along the gravity axis on four EMDB-2 sequences, each shown with one input frame.
  Ground truth (gray band), the GVHMR rollout without GravHMR (orange), and GravHMR (blue).
  Each prediction is aligned to the ground truth by one Sim(3) fitted on the first 100 frames, and the whole sequence is drawn.
  Dashed segments connect the rollout and the ground truth at the same instant.
  Numbers at the top right are the mean horizontal root error over the sequence.}
  \label{fig:traj_topdown}
\end{figure*}

\paragraph{Global trajectories.}
Fig.~\ref{fig:traj_topdown} shows whole-sequence top-down
trajectories of the GVHMR backbone with and without GravHMR,
complementing the windowed renderings of
Fig.~\ref{fig:supp_wwo_gravhmr}.
Without GravHMR, the rollout drifts on the stair and
rectangle walks and barely moves on the skateboard sequence;
GravHMR stays within 0.2--0.3\,m of the reference on the walks.

\begin{figure*}[!tp]
  \centering
  \includegraphics[width=\textwidth]{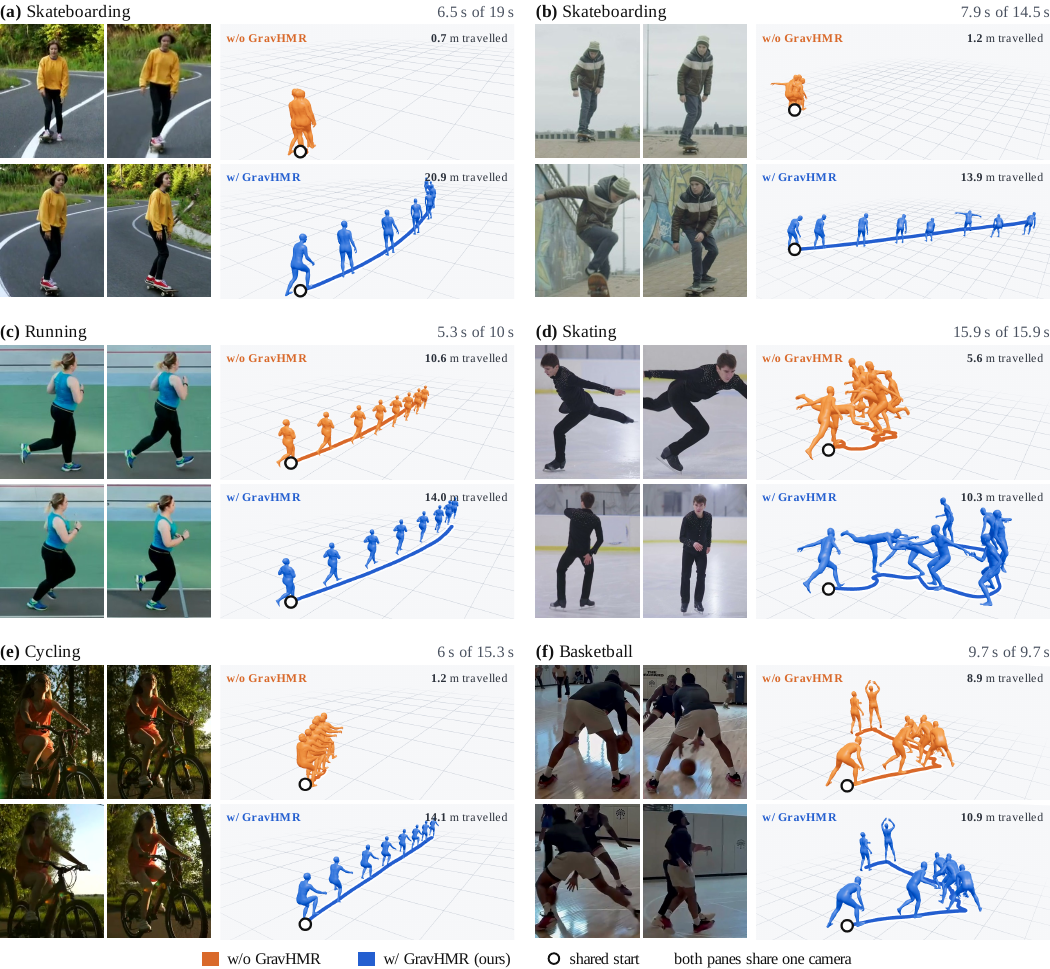}
  \caption{\textbf{In-the-wild videos, with and without GravHMR.}
  Six clips for which no ground truth exists. Each cell shows four input frames, then the
  GVHMR\,+\,PVA-Net rollout (\textcolor{rolloutorange}{orange}, top) and GravHMR
  (\textcolor{figblue}{blue}, bottom) over the same window, at eight time steps each;
  the circle marks the start the two share.
  \emph{One camera is fitted to both reconstructions and used for both panes}, so the
  difference in travelled distance is measured rather than produced by the viewpoint.
  A pane cannot hold a long trajectory and still show a body, so each clip is drawn from its
  start until GravHMR has travelled $14$\,m, or for a third of the clip, whichever is longer;
  the header gives the window and the full clip length, and the distances are measured inside
  the window. Cameras are from ViPE and the world is set upright with GeoCalib gravity;
  the rollout is already expressed in a gravity-aligned frame and is left untouched.}
  \label{fig:supp_wild}
\end{figure*}

\paragraph{In-the-wild videos.}
Fig.~\ref{fig:supp_wild} applies GravHMR to six in-the-wild
videos of running, skateboarding, skating, cycling, and
basketball with ViPE cameras and GeoCalib gravity, and puts each
beside the GVHMR\,+\,PVA-Net rollout it starts from. These videos
have no ground truth, so the comparison is qualitative: what it
shows is that the rollout stays near where it began while
GravHMR follows the motion out across the scene.

\begin{figure*}[!tp]
  \centering
  \includegraphics[width=\textwidth]{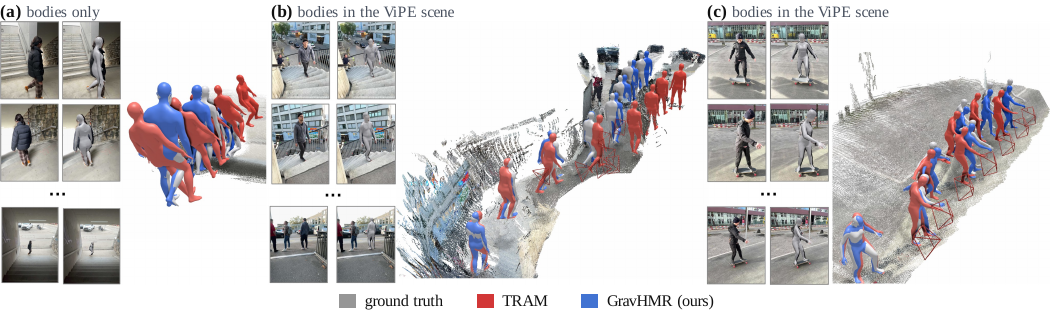}
  \caption{\textbf{Global human motion placed in the scene.}
  Ground truth (\textcolor{gtgray}{gray}), TRAM (\textcolor{tramred}{red}) and
  GravHMR (ours, \textcolor{oursblue}{blue}) at several time steps, beside sampled
  input frames with ground-truth overlays.
  The scenes in (b) and (c) are ViPE's own reconstruction, carried over unchanged:
  GravHMR never reads them, and nothing here is optimized against them.
  We show them because the bodies land in the scene the camera reconstructed
  without being asked to, which is what a future joint optimization of body and
  scene would start from. Panel (a) carries no scene and shows the bodies alone.}
  \label{fig:qualitative_global}
\end{figure*}

\paragraph{Body--scene consistency.}
Fig.~\ref{fig:qualitative_global} shows GravHMR, TRAM, and
ground-truth bodies at several time steps, overlaid on the
scene reconstruction produced by the SLAM stage.
GravHMR never reads the scene and nothing here is optimized
against it; the bodies land in the scene recovered by ViPE,
which is a natural starting point for a joint body--scene
optimization left to future work.

\paragraph{Robustness to broadcast-camera zoom.}
Tab.~\ref{tab:why-not-transloc} evaluates world-frame
trajectory recovery on WorldPose, captured under broadcast
camera zoom. A TRAM-style translocation does not recover usable
trajectories in this setting, whereas GravHMR remains stable.

\begin{table*}[!tp]
\centering
\caption{\textbf{Camera-space reconstruction on the RICH test set.}
PA-MPJPE, MPJPE, and PVE are in mm, and
Accel is in m/s$^2$.
Baseline results are taken from GVHMR~\cite{shen2024world},
while ours is evaluated on 191 views following
the GVHMR metric protocol.
Differences in inference settings may remain.
Best in \textbf{bold}, second best \underline{underlined}.}
\label{tab:rich_cam}
\setlength{\tabcolsep}{4.5pt}
\renewcommand{\arraystretch}{1.08}
\begin{tabular}{lcccc}
\toprule
Method
& PA-MPJPE $\downarrow$
& MPJPE $\downarrow$
& PVE $\downarrow$
& Accel $\downarrow$ \\
\midrule
\rowcolor{gray!15}
\multicolumn{5}{l}{\textit{Per-frame methods}} \\
SPIN~\cite{kolotouros2019learning}
& \rgbrank{69.7}{0}  & \rgbrank{122.9}{0}  & \rgbrank{144.2}{0}  & \rgbrank{35.2}{0} \\
CLIFF~\cite{li2022cliff}
& \rgbrank{56.6}{25} & \rgbrank{102.6}{29} & \rgbrank{115.0}{29} & \rgbrank{22.4}{13} \\
HMR2.0~\cite{goel2023humans}
& \rgbrank{48.1}{50} & \rgbrank{96.0}{43}  & \rgbrank{110.9}{43} & \rgbrank{18.8}{25} \\
ReFit~\cite{ReFit2023}
& \rgbrank{47.9}{63} & \rgbrank{80.7}{57}  & \rgbrank{92.9}{57}  & \rgbrank{17.1}{38} \\
\midrule
\rowcolor{gray!15}
\multicolumn{5}{l}{\textit{Temporal methods}} \\
TCMR~\cite{choi2021beyond}
& \rgbrank{65.6}{13} & \rgbrank{119.1}{14} & \rgbrank{137.7}{14} & \rgbrank{5.0}{75} \\
SLAHMR~\cite{ye2023decoupling}
& \rgbrank{52.5}{38} & --                   & --                   & \rgbrank{9.4}{50} \\
WHAM~\cite{shin2024wham}
& \rgbrank{44.3}{75} & \rgbrank{80.0}{71}  & \rgbrank{91.2}{71}  & \rgbrank{5.3}{63} \\
GVHMR~\cite{shen2024world}
& \rgbrankU{39.5}{88} & \rgbrankU{66.0}{86}  & \rgbrankU{74.4}{86}  & \rgbrankU{4.1}{88} \\
\midrule
\textbf{GravHMR (Ours)}
& \rgbrankB{35.5}{100} & \rgbrankB{62.8}{100} & \rgbrankB{70.7}{100} & \rgbrankB{3.35}{100} \\
\bottomrule
\end{tabular}
\end{table*}

\begin{table*}[!tp]
\centering
\caption{\textbf{Component swap between the GVHMR rollout and GravHMR.}
\ding{51} takes the part from the full core (before FATC), \ding{55}
from the GVHMR rollout; the trajectory is re-integrated without
re-optimization. \emph{Heading} is the body and travel yaw (YAOO);
\emph{speed \& vertical} is the horizontal speed and vertical motion
of the root (GATO). Body pose remains unchanged.
W-RVE is evaluated after a single yaw alignment over the first 2\,s.
Best results are marked in bold.}
\label{tab:ablation_yaoo_GATO}
\setlength{\tabcolsep}{4pt}
\renewcommand{\arraystretch}{1.10}
\begin{tabular}{ccccccc}
\toprule
\multicolumn{2}{c}{Part from GravHMR}
& \multicolumn{2}{c}{Global motion}
& \multicolumn{3}{c}{Root trajectory} \\
\cmidrule(lr){1-2}\cmidrule(lr){3-4}\cmidrule(lr){5-7}
Heading & Speed \& vertical
& W-MPJPE$_{100}\downarrow$
& WA-MPJPE$_{100}\downarrow$
& RTE$\downarrow$
& ERVE$\downarrow$
& W-RVE$\downarrow$ \\
\midrule
\rowcolor{gray!15}
\multicolumn{7}{l}{\textit{EMDB-2}} \\
\addlinespace[2pt]
\ding{55} & \ding{55} & 262.8 & 106.7 & 2.10 & 11.6 & 8.13 \\
\ding{51} & \ding{55} & 247.4 & 97.8  & 1.88 & 11.2 & 7.28 \\
\ding{55} & \ding{51} & 245.3 & 91.0  & 2.09 & 12.1 & 8.03 \\
\ding{51} & \ding{51}
& \textbf{155.3} & \textbf{57.6} & \textbf{1.27}
& \textbf{10.2} & \textbf{5.81} \\
\midrule
\rowcolor{gray!15}
\multicolumn{7}{l}{\textit{EgoBody}} \\
\addlinespace[2pt]
\ding{55} & \ding{55} & 203.5 & 129.4 & 3.85 & 9.4 & 6.58 \\
\ding{51} & \ding{55} & 167.2 & 111.6 & 3.02 & \textbf{8.8} & 5.53 \\
\ding{55} & \ding{51} & 234.7 & 128.7 & 6.40 & 10.3 & 6.70 \\
\ding{51} & \ding{51}
& \textbf{123.3} & \textbf{87.5} & \textbf{1.04}
& 8.9 & \textbf{3.88} \\
\bottomrule
\end{tabular}
\end{table*}

\begin{table*}[!tp]
\centering
\caption{\textbf{Effect of the camera intrinsics on EMDB-2 (25 sequences, frame-weighted).}
$|\Delta f|$ is the mean absolute focal-length error relative to ground truth.
TRAM uses the named intrinsics for both masked DROID-SLAM and its human regressor, so its camera poses differ between the two rows.
Ours (ViPE, GT K) keeps the ViPE poses and uses ground-truth intrinsics in the HMR input, GATO, and FATC.
The last two rows use masked DROID-SLAM, as \emph{Ours w/ masked DROID-SLAM} in Tab.~\ref{tab:world}.
WA$_{100}$, W$_{100}$, PA-MPJPE, and FS are in mm, RTE in \%, Jitter in $10\,\mathrm{m/s^3}$.
All rows are computed with one evaluator, so TRAM can differ from the values reported in Tab.~\ref{tab:world}.}
\label{tab:emdb2_intrinsic}
\setlength{\tabcolsep}{5pt}
\begin{tabular}{lll c ccc ccc c}
\toprule
Method & Poses & $\mathbf{K}$ & $|\Delta f|$ (\%)
& PA-MPJPE & WA$_{100}$ & W$_{100}$ & RTE & ERVE & Jitter & FS \\
\midrule
TRAM & masked DROID & TRAM est. & 7.7 & 37.9 & 104.9 & 304.9 & 2.97 & 20.2 & 106.1 & 19.5 \\
TRAM & masked DROID & GT & -- & 37.7 & 80.1 & 222.2 & 1.42 & 18.9 & 104.6 & 18.2 \\
\midrule
Ours & ViPE & ViPE est. & 0.6 & 29.5 & 57.4 & 154.0 & 1.29 & 9.2 & 8.4 & 5.2 \\
Ours & ViPE & GT & -- & 29.5 & 57.4 & 154.1 & 1.29 & 9.2 & 8.4 & 5.2 \\
\midrule
Ours & masked DROID & TRAM est. & 7.7 & 29.7 & 92.3 & 270.4 & 2.99 & 10.7 & 6.5 & 7.1 \\
Ours & masked DROID & GT & -- & 29.5 & 66.6 & 189.8 & 1.43 & 9.2 & 6.7 & 5.4 \\
\bottomrule
\end{tabular}
\end{table*}

\begin{table*}[!tp]
\centering
\caption{\textbf{Gravity source on EMDB-2.}
The body orientation $\mathbf{R}_t$ is rotated minimally so that its vertical
matches the named source; in the main configuration it is the vertical of the
HMR rollout (Eq.~\eqref{eq:yaoo_apply}). In all rows the world frame is set by
GeoCalib (Eq.~\eqref{eq:gravity_axis}), the cameras are from ViPE, and results
are before FATC. Units follow Tab.~\ref{tab:world}.}
\label{tab:gravity_oracle}
\setlength{\tabcolsep}{4pt}
\begin{tabular}{lccccc}
\toprule
Gravity source & WA$_{100}$ & W$_{100}$ & ERVE & Jitter & FS \\
\midrule
HMR rollout (main results) & 57.6 & 155.3 & 10.2 & 32.1 & 7.8 \\
GeoCalib & 62.8 & 155.3 & 10.7 & 36.6 & 8.0 \\
Ground truth & 63.5 & 160.0 & 10.8 & 35.8 & 6.7 \\
\bottomrule
\end{tabular}
\end{table*}

\begin{table*}[!tp]
\centering
\caption{\textbf{Runtime per stage} on three EMDB-2 sequences (6170 frames in total).
HMR, PVA-Net, GATO, and FATC were measured on a Tesla V100.
HMR2 and GVHMR times exclude model loading and image decoding; the FATC time includes start-up and post-processing.
GATO and the FATC optimization use a fixed iteration count, so they take about 10\,s and 80\,s per sequence regardless of length.
ViPE and GeoCalib times come from earlier run logs; ViPE ran at $384\times512$ on most sequences.}
\label{tab:runtime}
\setlength{\tabcolsep}{5pt}
\begin{tabular}{lrr}
\toprule
Stage & Seconds & FPS \\
\midrule
ViPE (full) & 3233 & 1.9 \\
GeoCalib (1 in 5 frames) & 267 & 23.1 \\
HMR2 features & 253 & 24.4 \\
GVHMR forward & 5 & 1190 \\
PVA-Net crops (CPU) & 357 & 17.3 \\
PVA-Net inference & 323 & 19.1 \\
\textbf{YAOO + GATO} & \textbf{28} & \textbf{220} \\
FATC optimization & 248 & 24.9 \\
\bottomrule
\end{tabular}
\end{table*}

\begin{table*}[!tp]
\centering
\caption{\textbf{World-frame trajectory recovery on WorldPose.}
TRAM-style translocation fails under broadcast camera zoom,
while ours remains robust.}
\label{tab:why-not-transloc}
\begin{tabular}{lccc}
\toprule
& W-MPJPE$\downarrow$ & WA-MPJPE$\downarrow$ & RTE(\%)$\downarrow$ \\
\midrule
TRAM-style translocation & 2651.1 & 955.0 & 8.8 \\
Ours w/o optimization    & 415.3  & 94.8  & 1.4 \\
\textbf{Ours}            & \textbf{405.4} & \textbf{93.9} & \textbf{1.1} \\
\bottomrule
\end{tabular}
\end{table*}

\begin{table*}[!tp]
\centering
\begin{tabular}{cc cc cc}
\toprule
\multirow{2}{*}{Structural} &
\multirow{2}{*}{Dynamic} &
\multicolumn{2}{c}{EMDB-2} &
\multicolumn{2}{c}{EgoBody} \\
\cmidrule(lr){3-4}\cmidrule(lr){5-6}
& & WA-MPJPE$\downarrow$ & W-MPJPE$\downarrow$
  & WA-MPJPE$\downarrow$ & W-MPJPE$\downarrow$ \\
\midrule
            &            & 58.5 & 156.7 & 149.1 & 232.3 \\
\checkmark  &            & 58.0 & 156.0 & 122.9 & 181.9 \\
            & \checkmark & 57.7 & 155.5 & 102.1 & 144.9 \\
\checkmark  & \checkmark & 58.0 & 155.6 & \textbf{97.2} & \textbf{136.1} \\
\midrule
\multicolumn{2}{c}{Detector Confidence}
            & 57.7 & 155.3 & 133.0 & 205.0 \\
\bottomrule
\end{tabular}
\caption{\textbf{Ablation of per-joint weighting strategies.}
Comparison of structural and dynamic weighting in GATO on
EMDB-2 and EgoBody, with the depth of the per-frame 2D root solve
(\emph{2D body extent} in Tab.~\ref{tab:depth_src}). Both strategies
improve robustness under challenging body observations.}
\label{tab:joint_weight}
\end{table*}

\end{document}